\pdfoutput=1

\documentclass[11pt]{article}

\usepackage{ACL2023}

\usepackage{times}
\usepackage{latexsym}

\usepackage[T1]{fontenc}

\usepackage[utf8]{inputenc}

\usepackage{microtype}

\usepackage{inconsolata}

\usepackage{amsmath}
\usepackage{cleveref}
\usepackage{hyperref}
\usepackage{booktabs}
\usepackage{graphicx}
\usepackage{algorithm}
\usepackage{algpseudocode}
\usepackage{soul}
\usepackage{xspace}
\usepackage{enumitem}
\usepackage{xcolor}
\usepackage{array}
\usepackage[skins]{tcolorbox}
\tcbuselibrary{listingsutf8}
\usepackage[labelformat=simple]{subcaption}

\usepackage{annotation}

\newcounter{question_ctr}
\newcommand{\question}[1]{%
\refstepcounter{question_ctr}%
\noindent\textbf{Q\arabic{question_ctr}}. \textit{#1} \\[3pt]}

\global

\tcbset{
  aibox/.style={
    width=474.18663pt,
    top=10pt,
    colback=white,
    colframe=black,
    colbacktitle=black,
    enhanced,
    center,
    attach boxed title to top left={yshift=-0.1in,xshift=0.15in},
    boxed title style={boxrule=0pt,colframe=white,},
  }
}
\newtcolorbox{AIbox}[2][]{aibox,title=#2,#1}

\newcolumntype{H}{>{\setbox0=\hbox\bgroup}c<{\egroup}@{}}

\definecolor{aigold}{RGB}{244,210, 1} 
\definecolor{aigreen}{RGB}{210,244,211} 
\definecolor{aired}{RGB}{255,180,181} 
\definecolor{forestgreen}{RGB}{0,120,90} 
\definecolor{lightblue}{HTML}{E4F6FF} 

\DeclareRobustCommand{\hlgold}[1]{{\sethlcolor{aigold}\hl{#1}}}
\DeclareRobustCommand{\hlgreen}[1]{{\sethlcolor{aigreen}\hl{#1}}}
\DeclareRobustCommand{\hlred}[1]{{\sethlcolor{aired}\hl{#1}}}
\DeclareRobustCommand{\hllightblue}[1]{{\sethlcolor{lightblue}\hl{#1}}}

\title{
    \vspace*{-0.5in}
    {\raggedright\small EMNLP 2024\\[0.25in]}
    Paraphrase Types Elicit Prompt Engineering Capabilities
}

\author{Jan Philip Wahle$^{*1}$, Terry Ruas${^1}$, Yang Xu${^2}$, Bela Gipp${^1}$\\
${^1}$University of Göttingen, Germany\\
${^2}$University of Toronto, Canada\\
${^*}$\texttt{wahle@uni-goettingen.de}\\}

\begin{document}
\maketitle
\AddAnnotationRef

\begin{abstract}
Much of the success of modern language models depends on finding a suitable prompt to instruct the model. Until now, it has been largely unknown how variations in the linguistic expression of prompts affect these models. This study systematically and empirically evaluates which linguistic features influence models through paraphrase types, i.e., different linguistic changes at particular positions. We measure behavioral changes for five models across 120 tasks and six families of paraphrases (i.e., morphology, syntax, lexicon, lexico-syntax, discourse, and others). We also control for other prompt engineering factors (e.g., prompt length, lexical diversity, and proximity to training data). Our results show a potential for language models to improve tasks when their prompts are adapted in specific paraphrase types (e.g., 6.7\% median gain in Mixtral 8x7B; 5.5\% in LLaMA 3 8B; cf. \Cref{fig:teaser}). In particular, changes in morphology and lexicon, i.e., the vocabulary used, showed promise in improving prompts. These findings contribute to developing more robust language models capable of handling variability in linguistic expression. Code: \url{https://github.com/jpwahle/emnlp24-prompt-paraphrase}.
\end{abstract}

\section{Introduction}

\begingroup
\addtolength\leftmargini{-0.3cm}
\begin{quote}
    \textit{It's not what you say it's how you say it.}\\[5pt]
    \hspace*{\fill} --- Albert Mehrabian 
\end{quote}
\endgroup

\noindent

\noindent Large language models (LLMs) already mimic human interaction by receiving instructions through natural language prompts and responding in natural language \cite{radford2019language, ouyang2022training, touvron2023llama}.
The way prompts are designed has a marked impact on the value of a model's output, and current LLMs require some degree of prompt engineering to be successful \cite{LIU2023, lu2023prompts, leidinger2023language}. 
A key step in prompt engineering is understanding how humans can express similar meanings differently, also known as paraphrasing. 
For example, consider the following paraphrases of a prompt, which share no words and vary greatly in length but convey the same message:

\begin{quote}
\textit{Avoid procrastination.\\Stop postponing what you have to do.}
\end{quote}

\begin{figure}[t]
    \centering
    \includegraphics[width=\columnwidth]{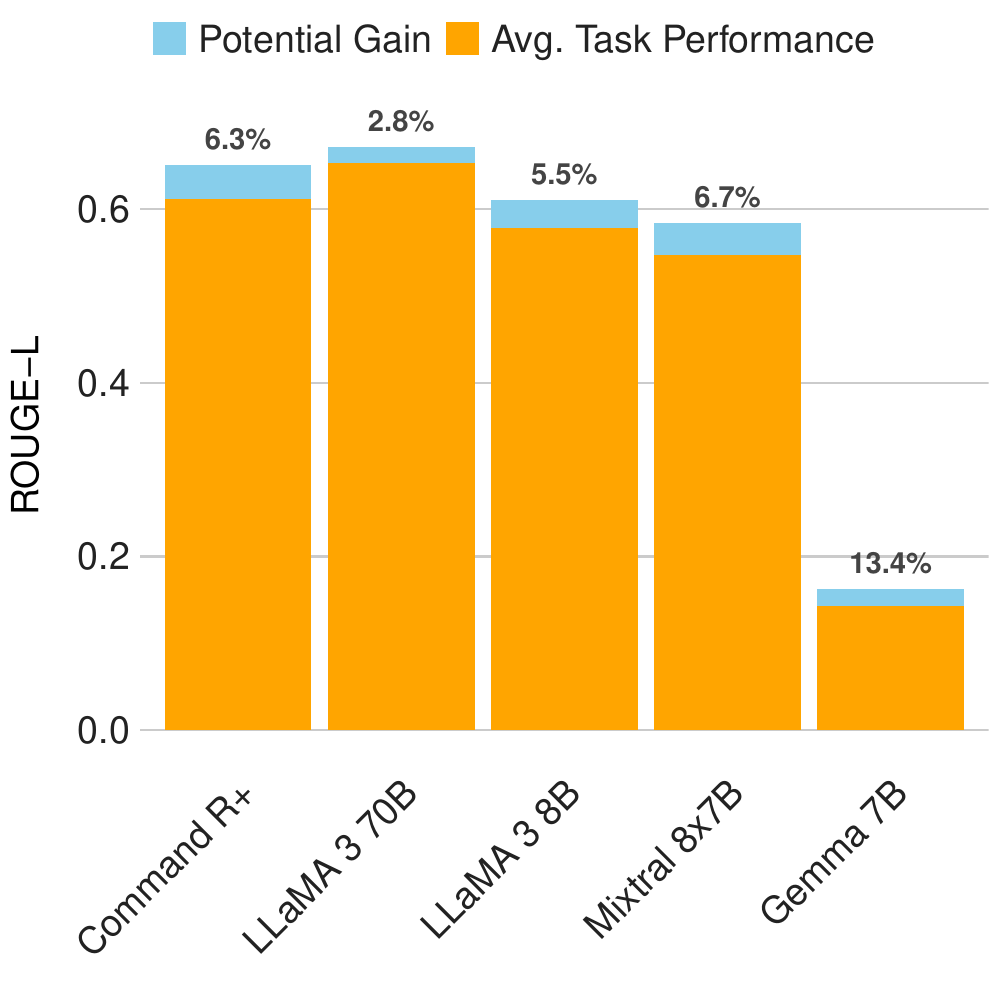}
    \caption{The potential median task performance gain (blue) over the model's baseline performance (orange) of five chat models across 120 tasks when their prompts were adjusted for specific paraphrase types (e.g., lexicon, syntax, morphology).}
    \label{fig:teaser}
\end{figure}

\begin{figure*}
    \centering
    \includegraphics[width=\textwidth]{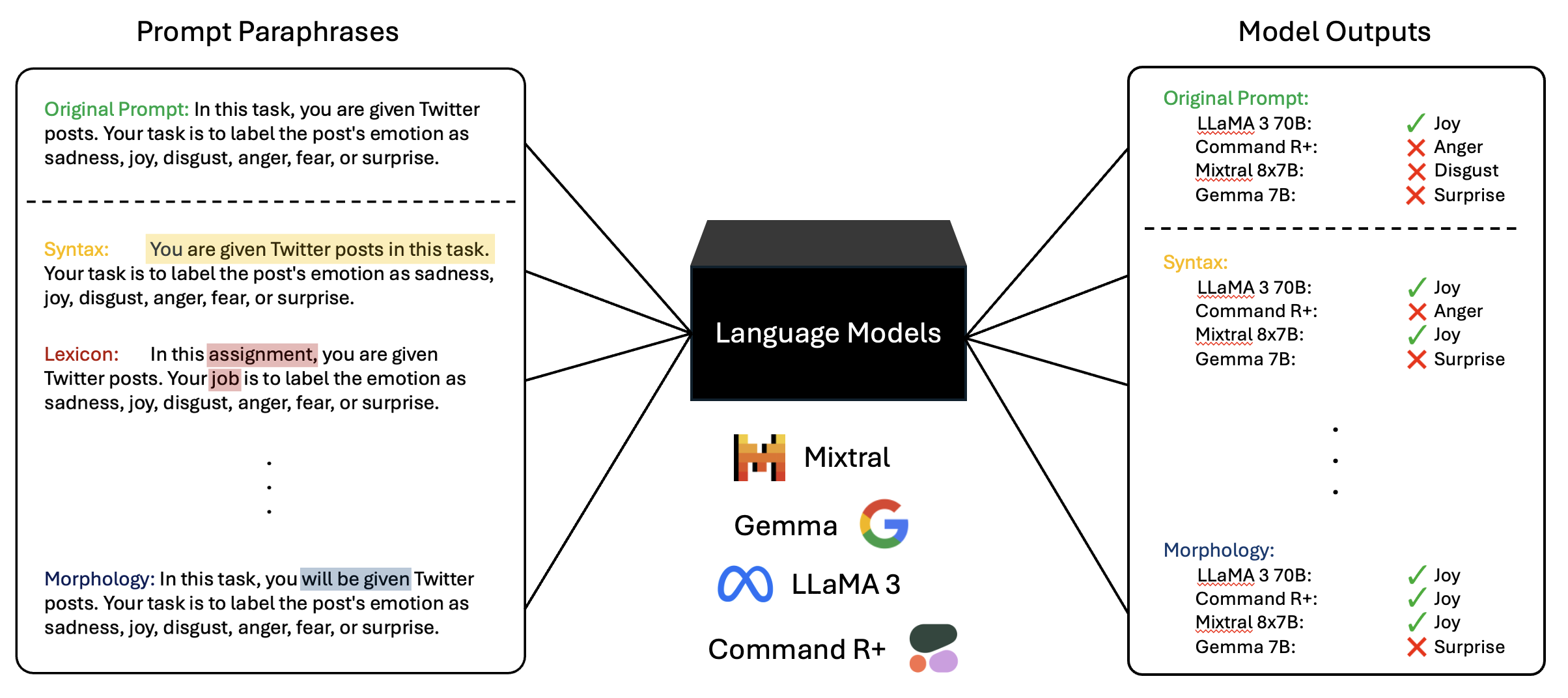}
    \caption{The main method of this paper. We paraphrase prompts of 120 tasks from 24 task families using 26 linguistic types of six categories (i.e., morphology, syntax, lexicon, lexico-syntax, discourse, and others) to analyze model inputs and outputs across different factors.}
    \label{fig:teaser_method}
\end{figure*}

Humans understand and interpret the diversity in expressions, often without conscious effort.
Arguably, LLMs should handle linguistic flexibility in a similar way to humans.
Paraphrasing provides a window into the heart of prompt engineering; it gives us insight into what characteristics of the instructions language models value, what they understand, and where they lack capabilities.
Ideally, LLMs should be robust to lexical, syntactic, morphological, inter alia, changes in the provided instruction --- similar to how humans understand semantically identical texts written differently.  

Through paraphrases, we can adjust prompts in these specific types to better understand which influence LLM's ability to solve tasks. 
Emerging tracks in conferences and workshops on how to design prompts underline a substantial interest in understanding which nuances of prompts affect models \cite{promptengworkshop}.

To determine the influence of linguistic types on prompts, we need a shared understanding of the variations between similar prompts. 
\citet{Vila2014} introduced \textit{paraphrase types} which define linguistic changes at individual text positions. 
\citet{wahle-etal-2023-paraphrase} adopted paraphrase types to train models that generate paraphrases with these capabilities. 
Paraphrase types can identify linguistic types, such as whether the lexicon has changed (by replacing words) or the syntax has changed (by altering grammatical structure).

So far, there has been little work to systematically evaluate how linguistic variations of prompts affect current language models.
We fill this gap by evaluating hundreds of thousands of prompt variations across 120 tasks and five models (3.24m total prompt-example combinations). 
Our procedure is outlined in \Cref{fig:teaser_method}.
We measure how sensitive the models are to different prompt changes and what we need to improve their robustness and effectiveness in performing different tasks. 
Our work addresses the following key questions:
\begin{enumerate}
    \itemsep0.2em 
    \item How sensitive are the models to specific paraphrase perturbations of the prompt?
    \item Which linguistic changes in the prompt affect models the most?
    \item What other factors play a role, such as the length and lexical diversity of the prompt or proximity to the training data?
    \item How does the above evolve across models and tasks?
\end{enumerate}

Our results show that adapting prompts to specific types (e.g., morphology) can yield significant gains in many models (cf. \Cref{fig:teaser}) across different downstream tasks, such as summarization, text completion, or dialogue generation. 
We also control for various confounding factors, such as the length of prompts, lexical diversity, and proximity to training data, and show that performance gains can be observed independent of these additional factors. 
We recommend which paraphrase types to consider when adapting prompts for a particular task (e.g., polarity substitutions for sentiment analysis). 

\section{Related Work}

Prompt-based learning has become a new trend, where pre-trained language models perform prediction using a template description of the intended task, and the model derives the necessary information without the need for gradient updates \cite{LiuYF2023}. 
Prompt-based learning has advantages over traditional supervised learning, as the same pre-trained LLM can perform different unseen tasks without fine-tuning.
Early prompt tuning methods focused on integrating trainable continuous prompt embeddings to perform various NLU tasks \cite{gu2021ppt,liu2021p,LIU2023}.
Discrete prompts representing actual tokens have become more popular than continuous embeddings in prompt engineering, arguably because they are more accessible and linguistically interpretable. 

Discrete methods used cloze-style phrases and differentiable prompts to improve few-shot learning capabilities \cite{schick2020exploiting, zhang2021differentiable}.
\citet{shin2020autoprompt} proposed AutoPrompt, which uses a masked language model to find variations of prompts and has been used for various downstream tasks such as plagiarism detection \cite{wahle-etal-2022-large}.
\citet{zhou2022large} introduced Automatic Prompt Engineer, which searches over a pool of prompt candidates proposed by a second LLM.
Other work focused on generating prompts for knowledge extraction from LLMs \cite{JiangXAN2020} or investigated instruction induction and model tuning with minimal human intervention \citet{honovich2022instruction,honovich2022unnatural}. Recent work uses variations of Self-Instruct to improve model prompting \cite{wang2022self}.

\citet{Reynolds2021, hu-levy-2023-prompting} evaluate the role of prompts in model control and the limitations of metalinguistic prompts for assessing linguistic knowledge in LLMs. 
Studies by \citet{leidinger2023language} and \citet{lu2023prompts} examined the influence of linguistic features of prompts on LLM performance with hand-selected prompt variations (e.g., tense, modality). \citet{mizrahi2024state} demonstrated through empirical analysis that LLM performance can vary based on how task instructions are phrased. \citet{sorensen2022information, yang2023improving} explored the optimization of prompt selection using unified scoring frameworks and unsupervised techniques based on mutual information.

Our work contributes to previous work in several key aspects.
While previous studies have explored hand-crafted linguistic features on LLM performance \cite{leidinger2023language,lu2023prompts}, we take a systematic approach by decomposing paraphrases into a set of six families of changes (i.e., syntax, lexicon, lexico-syntax, morphological, semantic, and others). 
Although other works have proposed strategies for prompt selection \cite{yang2023improving,sorensen2022information} and methods for enhancing prompt tuning and generation \cite{liu2021p,shin2020autoprompt,zhou2022large}, our research quantitatively measures the effects of linguistic paraphrase types on model responses in a bottom-up method to apply successful types on different tasks. 
Other work focuses on domain-specific problems with few tasks.
Our methodology spans five models, 120 tasks, and 24 task families. 
The main gap we address is the lack of empirical evaluation of how different linguistic manipulations of prompts affect LLM behavior and performance on a large scale. 
Our work controls for other confounding factors, such as prompt complexity, training data proximity, and lexical diversity. 

\section{Data \& Models}

Central to this study is a dataset of tasks, prompts, and variations of these prompts. 
We use the Super-NaturalInstructions dataset \cite{wang-etal-2022-super}, which contains more than 1\,600 tasks with their respective prompts. 
We describe in detail how we construct different variations of these prompts in the experiments (\Cref{sec:experiments}).
We sample 120 different tasks (each task has its own dataset) from 24 different task families (e.g., question answering, sentiment analysis) with the following conditions: the task family must have at least 10 tasks, the primary language is English, and each task has 200 examples to provide sufficient statistical power. 
We sample 5 tasks per domain, leaving us with 24 task families x 5 tasks per domain = 120 total datasets. 
See \Cref{ap:prompts_tasks} for the task details.

We choose the five best non-proprietary models according to the LMSYS chatbot arena leaderboard in descending order of performance as of 1 May 2024\footnote{\url{https://chat.lmsys.org/?leaderboard}}: LLaMA 3 Instruct (70B), Command R+ (104B), Mixtral 8x7B Instruct (47B), LLaMA 3 Instruct (8B), and Gemma Instruct (7B). 
Unless otherwise noted, we set the temperature to 0.2, the probability mass to sample words (top p) to 0.9, the penalty for repeatedly sampling the same sequence of tokens to 0.1, and the maximum number of tokens generated to the average of human references for that task. 
This number varies for different tasks but averages between 1 and 3154 tokens (see \Cref{ap:prompts_tasks} for details). 
We use 40 NVIDIA A100 GPUs (40GB) with 16-bit precision (8 A100s for each model). 
Our experiments required a total computational budget of approximately 2880 A100 GPU hours, resulting in 311kg of CO2 equivalent\footnote{\url{https://mlco2.github.io/impact/\#compute}}. %

\section{Experiments}
\label{sec:experiments}

\noindent\question{How do different linguistic changes in the prompt affect the performance of a language model? Which changes affect a model the most?} 
\noindent\textbf{Ans.} We start with a seed prompt designed for a specific task, e.g., ``In this task, you are given Twitter posts. Your task is to label the post's emotion as sadness, joy, disgust, anger, fear, or surprise.''
We paraphrase each prompt with one of 26 linguistic changes (e.g., syntax, lexicon) using the gpt-3.5-turbo-16k model, which we fine-tuned in the same way as \citet{wahle-etal-2023-paraphrase} using the ETPC dataset \cite{kovatchev-etal-2018-etpc}.
See \Cref{tab:ap_types_by_group} and \Cref{fig:ap_prompt_template,fig:ap_prompt_example_numersense} in \Cref{ap:prompts_tasks} for an extensive list of types, full prompt template and examples respectively. 
For the original prompt and its 26 paraphrases, we test each model's few-shot settings. 
Depending on the task, the number of few-shot examples ranges from 3 to 6 (sum of positive and negative examples).

\begin{figure}[t]
    \centering
    \includegraphics[width=\columnwidth]{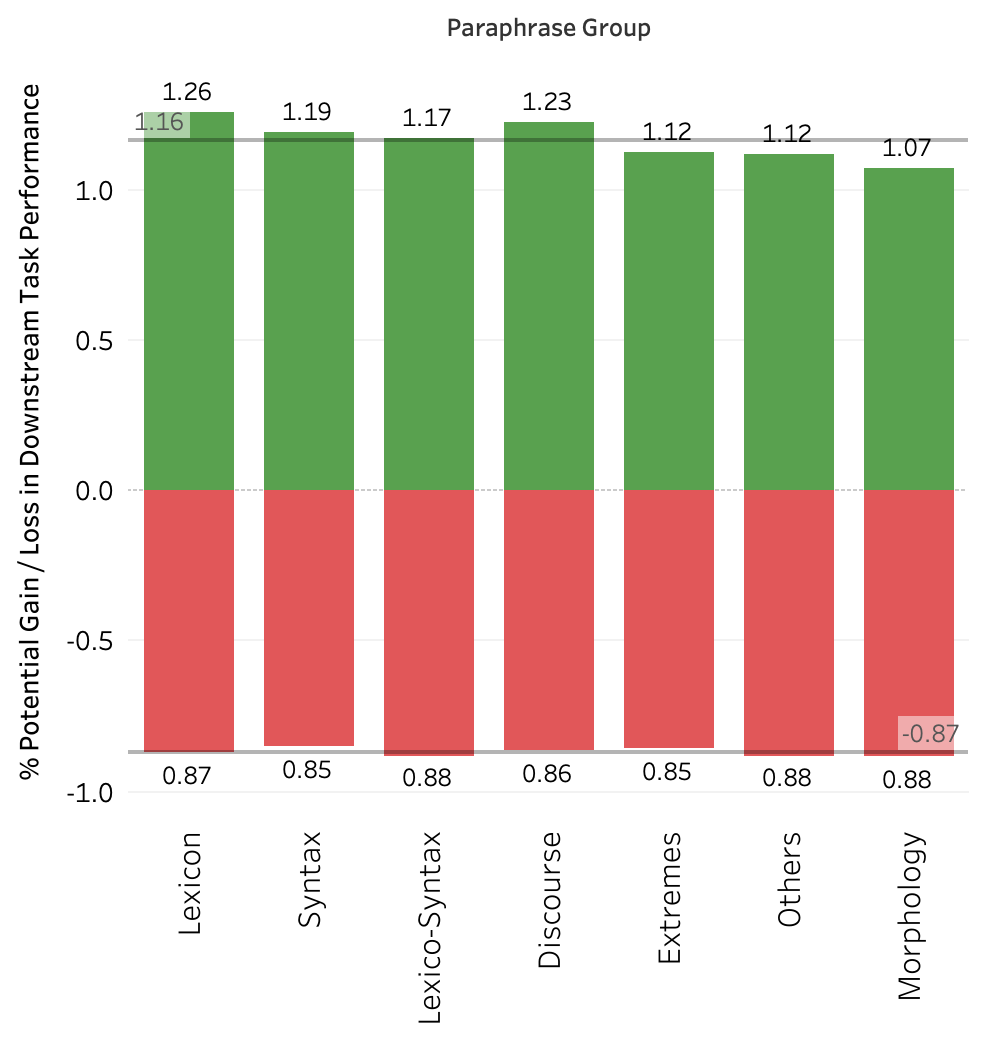}
    \caption{The average downstream task performance gain or loss from applying specific paraphrase types to the prompt for all 120 tasks and five models.}
    \label{fig:q1_dumbbell}
\end{figure}

\noindent\textbf{Results.} As \Cref{fig:teaser} already revealed, we can observe marked changes when adjusting prompts. 
Command R+ experiences up to 6.3\% median performance gain, and Gemma 7B up to 13.4\%. 
We decompose these results by the different types in \Cref{fig:q1_dumbbell}. 
Lexicon changes (+1.26\%), closely followed by syntax changes (+1.19\%), account for the largest median performance gain. 
The potential loss is comparable across changes. 
Overall, paraphrases had more upside than downside potential (+1.16\% vs. -0.87\% median change).

\noindent\textbf{Discussion.} Across all tested models, there seems to be marked upside potential when we adapt prompts in specific linguistic types. 
Morphology changes include changing modal verbs, which helps LLMs to follow instructions more clearly, e.g., ``In this task, one \st{should} \underline{must} detect the sentiment of the sentence.''
Lexicon changes have shown success across our experiments; in particular, we found examples in which more specific vocabulary benefits a task prompt, e.g., instead of ``Determine how people feel about this text.'' a more precise version yielded better results: ``Determine whether the sentiment expressed is positive, negative, or neutral'',

\question{Are there prompt changes consistently improving performance on a particular set of tasks? Do these changes have a different impact on tasks in different domains?}
\noindent\textbf{Ans.} We decompose the results of Q1 into individual task families, i.e., a set of related problems, such as sentiment analysis or question answering. 
A task is a specific set of data and instructions for solving a problem in a task domain, e.g., classifying emotions on X (formerly known as Twitter).

\begin{figure*}
    \centering  
    \includegraphics[width=\textwidth]{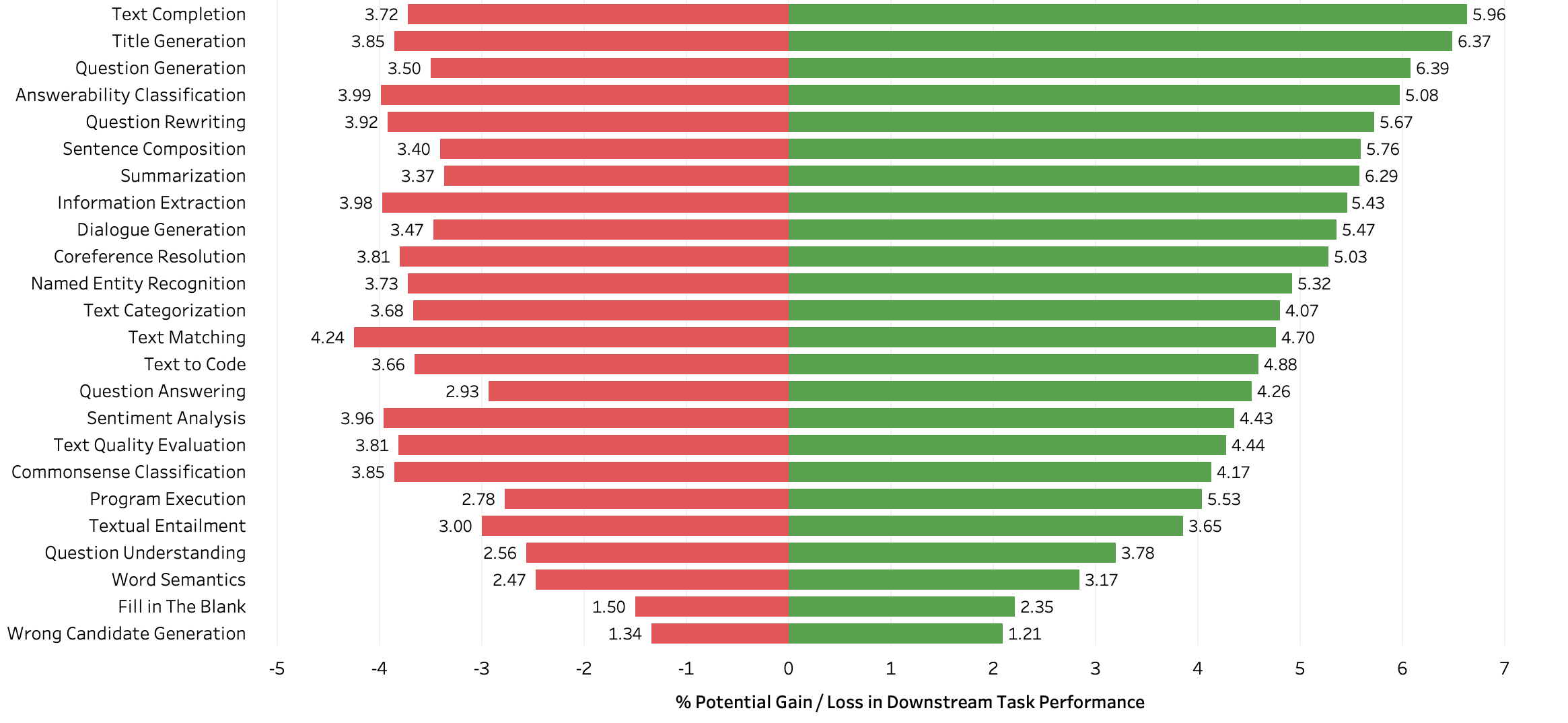}
    \caption{The avg. gain or loss in performance for all 120 tasks in 24 different task families across all five models.}
    \label{fig:q2_gain_loss_tasks}
\end{figure*}

\noindent\textbf{Results.} \Cref{fig:q2_gain_loss_tasks} shows that title generation (+6.01\%), text completion (+5.86\%), and question answering (+5.60\%) gain the most performance while having a low potential for a negative impact. %
Commonsense classification (-4.86\%), sentiment analysis (-4.83\%), and word semantics (-4.82\%) have the highest loss potential.

Specific perturbations of the prompt affect the performance of models on tasks from varying domains differently.
Morphology changes in the prompt have the largest gains in wrong candidate generation (+26.0\%), question generation (+21.5\%), and textual entailment (+17.5\%). 
Lexical changes show consistent gains in summarization (10.8\%), wrong candidate generation (+8.9\%), and title generation (+7.1\%). 
These tasks rely on semantic precision and vocabulary richness to interpret or generate nuanced responses. 
Positive effects with discourse changes can be observed in tasks that require an understanding of longer contexts or multi-sentence structures, such as summarization (+7.8\%).
Detailed results for each model are available in \Cref{tab:ap_task_by_types,tab:ap_task_by_types_gemma,tab:ap_task_by_types_command_r_plus,tab:ap_task_by_types_mixtral,tab:ap_task_by_types_llama_70b,tab:ap_task_by_types_llama_8b} in \Cref{ap:full_results}.

\noindent\textbf{Discussion.} Some tasks are more sensitive to instructions than others.
Tasks such as question generation and text completion may benefit from syntax changes because they can improve grammatical accuracy while reducing the flexibility of more creative tasks such as dialog generation. 
In particular, sentiment classification seems to benefit from polarity substitutions and is sensitive to negation, consistent with related work \cite{wiegand-etal-2010-survey}.
Lexical changes improve performance in vocabulary-intensive tasks such as named entity recognition and text entailment. 
However, they may hinder performance in more general contexts such as question rewriting, which is often based on data from Wikipedia and knowledge bases (see \Cref{tab:ap_overview_tasks_2} in \Cref{ap:prompts_tasks}).
In discourse, better context management, as well as indirect and direct style, improves clarity and coherence in tasks like summarization and long-form answer generation.

Sometimes, even the same changes can produce different results for different models.
For example, larger models such as Command R+ benefit from morphology changes for summarization (+23.6\%), while the same changes result in a loss of performance in the two small models LLaMA 3 8B (-16.4\%) and Gemma 7B (-16.1\%).

\noindent\question{How sensitive are different models and architectures to changes in the prompt? Are there differences between models of different sizes? Can a smaller model achieve better results than a larger model by paraphrasing its prompts?}
\noindent\textbf{Ans.} Related work has shown that model size and training scale play a central role in language model success, also known as ``scaling laws'' \cite{kaplan2020scaling, rae2021scaling, wei2022emergent}. 
What has not been shown so far is whether and how much lexical changes influence different model architectures and sizes when adapting prompts.
Specifically, we investigate how to bring the performance of a weaker model up to the level of a larger model by tweaking its prompts.

\noindent\textbf{Results.} In \Cref{fig:teaser}, note how the smaller 8B version of LLaMA 3 can gain up to 5.5\% median task performance, while the larger 70B model can gain much less with up to 2.8\%. 
LLaMA 3 8B has a lower baseline performance of 0.58, while the 70B model's baseline is at 0.65 (see \Cref{fig:q3_llama_scale_comparison} in \Cref{ap:full_results} for more details).

LLaMA 3 models are less sensitive to changes than other models (0.05 avg. std. in LLaMA models vs. 0.10 avg. std. over other models).
Smaller models are more sensitive and have the highest possible gain potential when adjusting prompts (e.g., Gemma 7B: +10.2\% gain through lexicon; LLaMA 8B +8.2\% gain through morphology).
The most sensitive model in our experiments is Command R+ (avg. std. 0.16).

Although Command R+ achieves an average score of 0.61, which is lower than LLaMA 3 70B's with 0.65, it can outperform LLaMA 3 70B by 0.08 when its prompts are tuned. 
The same applies to Mixtral 8x7B (0.55) and LLaMA 3 8B (0.58).
Mixtral's prompts, when tuned, can score 0.06 higher than the previously better LLaMA model.

If we always choose the best (paraphrased) prompts, we can achieve even higher performance gains; for example, LLaMA 3 8B could gain 21.1\%, making the model markedly better than its 70B counterpart by tweaking the prompts. 
We want to note that it is difficult to always find the best prompt. 
Therefore, we report median performance gains in the main body of the paper and report results for selecting the best prompt in \Cref{fig:teaser_max} in \Cref{ap:full_results}.

\noindent\textbf{Discussion.} Models have different architectures and training processes, especially training data, which play a fundamental role in their behavior. 
It is encouraging to see that there is still upside potential to improve task results without relying on more computational resources for training.
However, this raises the question of why such different behaviors can be achieved by changing the instructions.
An interesting parallel is how humans can also sometimes produce different results depending on the instructions given. 
Our results suggest that smaller models can perform similarly to larger ones and are more sensitive to paraphrase changes.

\noindent\question{Do paraphrased prompts that increase a model's task performance also show greater similarity to the model's training data?}
\noindent\textbf{Ans.} We know that a confounding factor in prompt engineering is that prompts that are closer to a model's training data improve their likelihood of answering more confidently \cite{zhao2021calibrate}. 
We use the FineWeb corpus with 350 billion tokens ($\approx$388GB) to search for examples close to paraphrased prompts and compute their similarity.

We build a BM25 index ($\approx$610GB), and for each prompt and its 26 variations, we query this index to find the closest examples. 
We measure the difference in similarity of the original prompt to the training data versus our paraphrased prompts to the training data by computing the following measure:

\begin{eqnarray}
    \Delta_{train} = RL(P, T_P) - RL(O, T_O)
\end{eqnarray}

\noindent where $RL$ is the ROUGE-L score, $O$ is the original prompt and $T_O$ is the nearest training example for $O$ (as measured by the BM25 score), $P$ is the paraphrased prompt, and $T_P$ is the closest training example for $P$.

\begin{figure}
    \centering
    \includegraphics[width=\columnwidth]{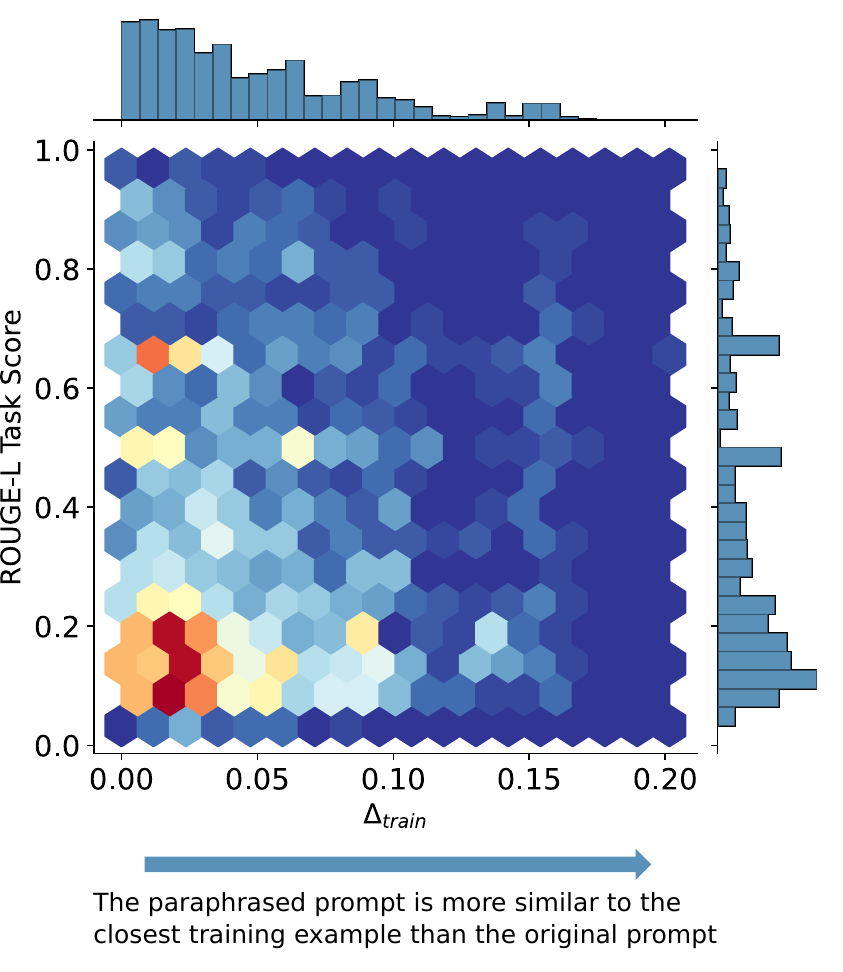}
    \caption{The distribution of how much closer the paraphrased prompt is to the closest training example in FineWeb 350BT over the original prompt (x-axis) and the distribution of task performance (y-axis). Red colors mean high mass and blue colors mean low mass between $\Delta_{train}$ and task performance.}
    \label{fig:closeness_training}
    \vspace*{-6mm}
\end{figure}

\noindent\textbf{Results.} %
The x-axis of \Cref{fig:closeness_training} shows $\Delta_{train}$, which represents this difference in similarity to the training data between the paraphrased and original prompts. 
The y-axis then shows downstream task performance. 
The results show few very close training examples for our tasks, i.e., the ROUGE-L between prompt and closest FineWeb example is typically below 0.5. 
Most successful paraphrased prompts with a downstream task performance greater than 0.8 do not show higher similarity to training than the original prompt. 
Sometimes, higher scores can be achieved when paraphrases are closer to the training (top right of the figure).

\noindent\textbf{Discussion.} Consistent with related work \cite{zhao2021calibrate}, we show that prompts closer to training examples sometimes have higher accuracy than those without close training examples. 
However, this does not generally hold across our large set of tasks. 
We found no evidence that paraphrases were better because they had closer training examples.

\begin{figure*}[t]
    \centering
    \includegraphics[width=\textwidth]{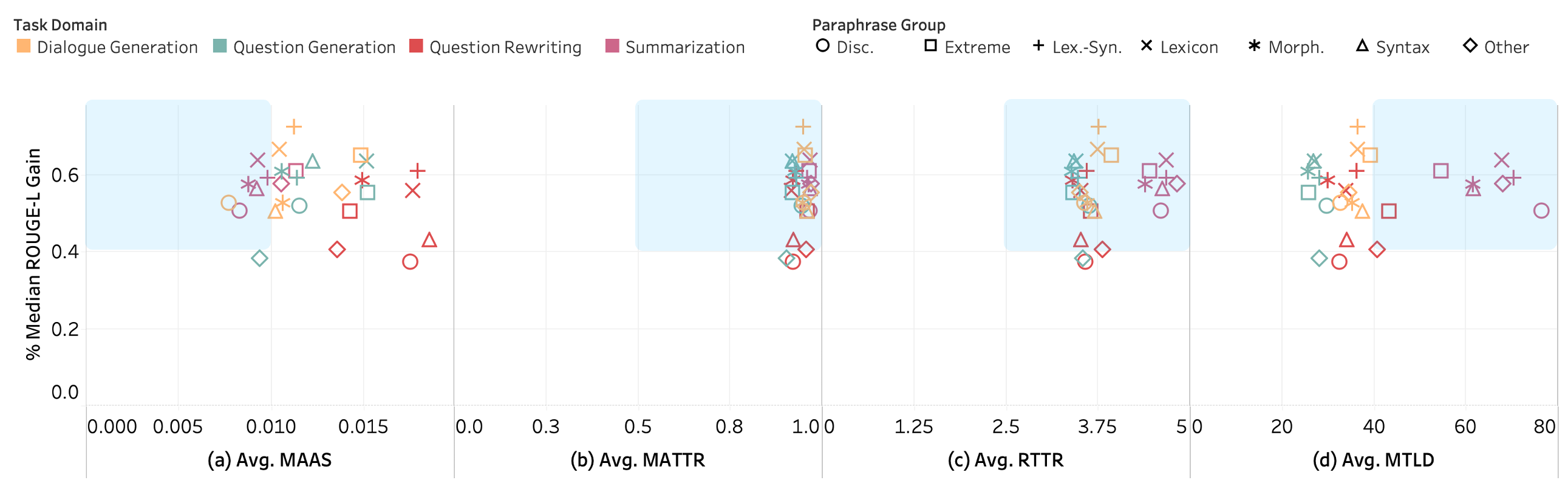}
    \caption{The percentage in task performance gain and lexical diversity as measured by four metrics in four tasks. The \hllightblue{light blue} quadrants show areas of high lexical diversity and strong task performance gains.}
    \label{fig:lexical_richness}
\end{figure*}

\noindent\question{How do different prompt perturbations affect the lexical richness of language model responses in generative tasks?} 
\noindent\textbf{Ans.}
We have previously measured downstream performance as a key indicator of prompt quality. 
In classification tasks, this is directly measurable by the binary downstream metric (i.e., ``yes'' or ``no'').
In generative tasks, it is not easy to say whether a generative answer is correct using a human reference \cite{clark-etal-2021-thats}; in particular, an output may be correct even without having any words in common with a reference (recall our example from the introduction). 
Therefore, we can evaluate complementary factors about responses (e.g., whether the response is creative or uses lexically diverse language).
A key factor in complex responses in natural text is lexical richness \cite{van2007comparing, jarvis2013capturing, kyle2019measuring}. 

We measure lexical richness with four metrics that focus on the diversity of tokens and text segments \cite{guiraud1960rttr,maas1972maas,covington2010cutting,mccarthy2010mtld}: Root Type-Token Ratio (RTTR), Maas, Moving Average Type-Token Ratio (MATTR), and Measure of Lexical Diversity (MTLD) --- more details in \Cref{ap:metrics}. 
We select four task families with long responses to measure lexical richness: summarization, question generation, question rewriting, and dialogue generation (20 tasks in total).

\noindent\textbf{Results.} Summarization tasks have a particularly high diversity in responses as measured by Maas, RTTR, and MTLD (\Cref{fig:lexical_richness} (a), (c), (d)). 
Overall, lexicon and lexico-syntactic changes often lead to higher performance at the expense of lexical richness. 
Models produce responses with lower lexical diversity in question rewriting tasks as measured by MAAS (\Cref{fig:lexical_richness} (a)). 
Syntax and other changes lead to both lower lexical diversity and task performance gains compared to others measured by RTTR (\Cref{fig:lexical_richness} (c)). 
Question rewriting achieves the highest lexical diversity as measured by MAAS, specifically with lexicon and lexico-syntactic changes (\Cref{fig:lexical_richness} (a)), and question generation and question rewriting have overall lower diversity and performance gains, specifically with discourse and other changes (\Cref{fig:lexical_richness} (c)). 

\noindent\textbf{Discussion.} Unexpectedly, summarization and question rewriting, which are arguably less open-ended and based more on extracting or paraphrasing information, lead to high lexical diversity. Purely generative and more open-ended tasks such as question generation and dialogue generation yield lower lexical diversity scores.
Larger changes in the prompt markedly affect the lexical diversity of model responses.
This is not surprising, as asking people to perform tasks in very different ways often leads to more diverse responses.
What is surprising, however, is that marked performance gains can still be observed.
Since these changes lead to substantial variance in the language model output, our results suggest that the complexity of the prompt (e.g., its length, word positions, or lexical shifts) can also be a confounding factor to task performance.

\noindent\question{What role does prompt complexity play in task performance outcomes? To what extent do variations in length, word position, and lexical variation correlate with downstream task performance?}
\noindent\textbf{Ans.} We use three markers to describe the complexity of a prompt paraphrase relative to the original instruction: the deviation in absolute number of tokens, word position deviation, and lexical deviation \cite{liu-soh-2022-towards}.
We use Pearson correlation to measure a quantitative relationship between these three markers of prompt complexity and task performance. 
More details about the metrics and correlations can be found in \Cref{ap:metrics}.

\begin{figure*}[!t]
    \includegraphics[width=\textwidth]{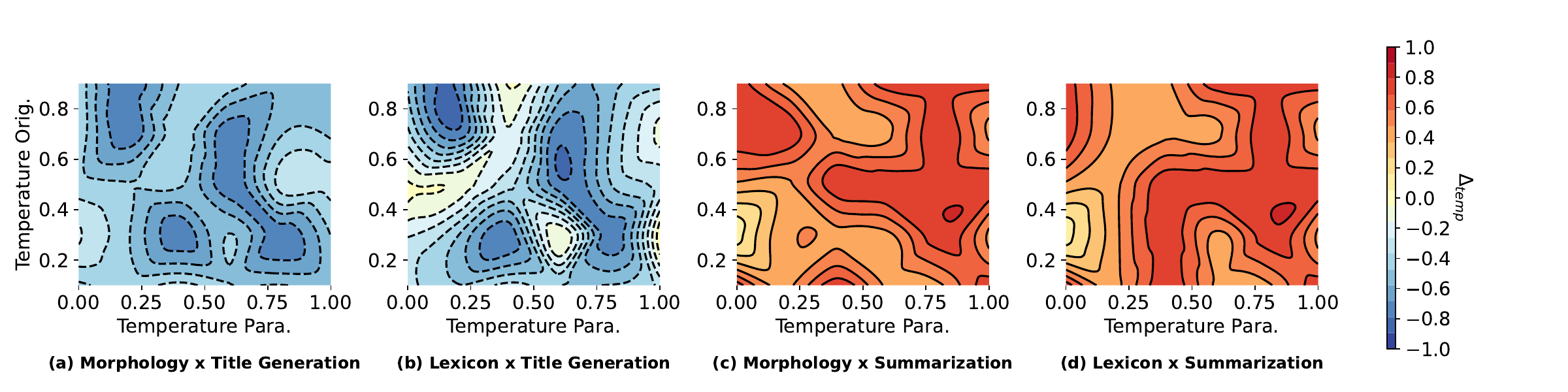}
    \caption{The performance difference between original and paraphrased prompts for temperatures from zero to one.}
    \label{fig:temperature}
\end{figure*}

\noindent\textbf{Results.} Across tasks, there is no significant correlation between changes in prompt length, changes in word position, and lexical deviation, as all correlations are close to zero (see \Cref{tab:prompt_complexity} in the \Cref{ap:full_results}). We found no evidence that making prompts more or less complex is associated with higher or lower task performance.

\noindent\textbf{Discussion.} Note that when paraphrasing prompts in certain lexical types, it is not new complexity added to the prompt that leads to different results (both gains and losses in task performance). 
This suggests that the actual linguistic perturbations make the prompt easier for the model to understand, leading to higher scores. 
However, there may be other factors that can also play a role.

\noindent\question{How much randomness is involved in these changes? Can the models' temperature lead to particularly good or bad task results that partially explain these gains or losses?}
\noindent\textbf{Ans.} When prompting models to perform tasks, we choose 11 temperature variations for LLaMA 3 70B from zero to one with steps of 0.1. 
For title generation and summarization, we run the original prompt and its paraphrased version using morphology and lexicon changes with these different temperature variations. 
We do this 11 times, with each run having a different temperature for the same paraphrased prompt, and then compute the difference between task performance of the original and paraphrased versions as:

\begin{eqnarray}
    \Delta_{temp} = RL(LM(O_i)) - RL(LM(P_j))
\end{eqnarray}

\noindent where $RL$ is the ROUGE-L score, $O_i$ is the original prompt with temperature $i$, $P$ is the paraphrased prompt with temperature $j$, and $LM$ is the language model output.

\noindent\textbf{Results.} \Cref{fig:temperature} shows contour plots of different temperature choices. 
The coloring shows $\Delta_{temp}$ for different temperatures for the paraphrased (x-axis) and original (y-axis) prompts. 
Note how the distributions within the summarization have similar hills and valleys (see \Cref{fig:temperature} (c) and (d)). 
For title generation, with generally higher performance loss, the highest gains are achieved for low to medium temperatures (up to 0.6 for the paraphrased prompt). 
For summarization, with overall higher scores, low temperature yields the highest performance difference, close to 0 for the paraphrased prompts and between 0.2 and 0.4 for the original.

\noindent\textbf{Discussion.} While we cannot fully exclude that randomness might improve results (i.e., there are some hills even in the high temperature ranges), most of the gains stem from low temperatures, suggesting that significant performance gains are due to paraphrase types. 
It may also be worth mentioning that additional randomness can come from the models trained by \cite{wahle-etal-2023-paraphrase}, as they can only paraphrase with some degree of accuracy.

\section{Conclusion}
This study evaluated five language models across 120 tasks (including sentiment, question answering, commonsense reasoning, summarization, etc.)  and showed that paraphrasing prompts can improve the performance of language models. 
We showed that language models have a marked upside potential to improve task results when their prompts are adapted in specific linguistic types (e.g., polarity substitutions for sentiment analysis).
Other work can use our findings about which tasks benefit from which paraphrase types to design new prompts. 
We also controlled for prompt complexity, temperature, and proximity to training data.

Current model performance represents a lower performance bound for tasks as we showed that semantically identical instructions hold marked upside gain. While it is not entirely clear why language models are often sensitive to changes in instruction, we have systematically tested different lexical features and found that some have a larger positive impact (e.g., morphology) than others (e.g., syntax), depending on the tasks. Since humans understand tasks presented in different ways and are robust to small (or even complex) changes in instruction, language models should have a similarly robust interface to communication in the future. We have contributed to this development of robust language interfaces by showing how specific types can benefit or harm models over a large set of tasks and prompts. One can also use our approach to create prompts the model does not understand to augment training data and increase its robustness.

\section*{Limitations}
Our results show that the same paraphrase changes can potentially improve results for one model but harm another.
The same is true for different changes in the same task. In a task like sentiment, sometimes the same polarity substitutions lead to improvements, and sometimes they do not.
Variance across examples and models does play a role here but another possible reason is that models trained to generate paraphrase types only have a certain accuracy, leading to models sometimes confusing one type for another or generating an incorrect type as recent work shows \cite{meier2024human}.
However, at a large aggregate level (3.24 million prompt-example combinations), our results provide important trends about prompts despite these margins of error. Better models for the controlled paraphrase generation will provide more accurate results in the future.

Our selection of paraphrase types is not a complete set of all flexibility in linguistic expression. There may be other variations, especially extremes, that we have not considered. However, the set of paraphrases across morphology, lexicon, lexico-syntax, syntax, and discourse covers many of the most common paraphrases encountered in texts.
We also find that there is no single paraphrase type that improves a model's accuracy across tasks consistently.
This is somewhat to be expected, as different tasks may benefit from different adaptations even though the overall domain is the same.
The same is true for tasks in different domains; some modifications are successful in one domain but not in another.
For example, polarity and negation play a larger role in sentiment, while lexical changes affect vocabulary intensive tasks (e.g., named entity recognition) or tasks that require specificity in vocabulary (e.g., example).
Finding the most successful type of change in a given setting is non-trivial, and more research needs to be done on successfully perturbed prompts for new and unseen tasks.

\section*{Acknowledgements}
This work was partially supported by the Lower Saxony Ministry of Science and Culture and the VW Foundation. Many thanks to Andreas Stephan and Lars Käsberg for their thoughtful discussions.

\bibliography{custom}
\bibliographystyle{acl_natbib}

\newpage

\appendix
\clearpage

\section{Details on the Prompts \& Tasks}
\label{ap:prompts_tasks}

\noindent\textbf{Prompts.} We use the template shown in \Cref{fig:ap_prompt_template} to construct prompts for each of the 120 tasks. In \Cref{fig:ap_prompt_example_numersense}, we show an example for the numerical commonsense reasoning task NumerSense in which the model has to predict a blank token `\textunderscore{}' \cite{lin2020birds}. To paraphrase prompts, we use one of 26 paraphrase types in six groups shown in \Cref{tab:ap_types_by_group}. Five examples of paraphrases applied to prompts can be seen in \Cref{tab:example_paraphrases}. In the first case, the added specification made the instruction more precise; in the second case, the model removed seemingly unnecessary context. The third case involved replacing words or phrases with synonyms or near-synonyms that have the same meaning in the given context, while the fourth case involved restructuring the sentence or changing the way the information is presented while maintaining the same overall meaning. Finally, the fifth case replaced a single word with a phrase (or vice versa) that conveys the same meaning. Further paraphrase type definitions and perturbation examples can be found in \cite{Vila2014,meier2024human}. 

\noindent\textbf{Tasks.} We provide an overview of all 120 tasks together with the paraphrase type that, when applied to prompts in that task domain, had the most positive influence together with the potential downstream task performance gain over the original prompt in \Cref{tab:ap_overview_tasks_1,tab:ap_overview_tasks_2,tab:ap_overview_tasks_3,tab:ap_overview_tasks_4}. Further, we want to note an observation of our experiments showing that tasks containing content judged as unethical (such as toxic language detection on Twitter) led to many denied responses by the chat or instruction-finetuned models. Likely because of their post-training, models answered that they could not respond to these questions, although we did not ask them to produce new harmful content but to classify existing content. We removed these tasks from our final analysis. This observation raises questions about how many powerful models can be used in the future to classify toxic content when their guardrails prevent them from answering.

\begin{figure*}[t]
    \begin{AIbox}{Task Prompt Template}
    \parbox[t]{\textwidth}{
            {\small \texttt{<system>}}\\
            \hspace*{5mm}{\small You are a helpful assistant in solving various tasks. You should only output one answer to the task, nothing more,\\
            \hspace*{5mm}no explanations, and nothing around. Just read the instruction carefully, understand the positive and negative\\
            \hspace*{5mm}examples provided, and generate one single answer in the same way as the example's output.}\\
            {\small \texttt{</system>}}
            
            \tcbline

            {\small \texttt{<user>}}\\
            \hspace*{5mm}{\small {\bf Instruction:} \{instruction\} }\\

            \hspace*{5mm}{\small {\bf Positive examples:} \{positive examples\} }\\
        
            \hspace*{5mm}{\small {\bf Negative examples:} \{negative examples\} }\\
        
            \hspace*{5mm}{\small {\bf Input:} \{example\} }\\
            \hspace*{5mm}{\small {\bf Output:}}\\
            {\small\texttt{</user>}}
    }
    \end{AIbox}
    \caption{The prompt template used to perform each of the 120 tasks. For Mixtral 8x7B Instruct and Gemma 7B Instruct, we prepend the system prompt to the user prompt because the models do not natively support system prompts.}
    \label{fig:ap_prompt_template}
\end{figure*}

\begin{figure}[t]
    \centering
    \includegraphics[width=\columnwidth]{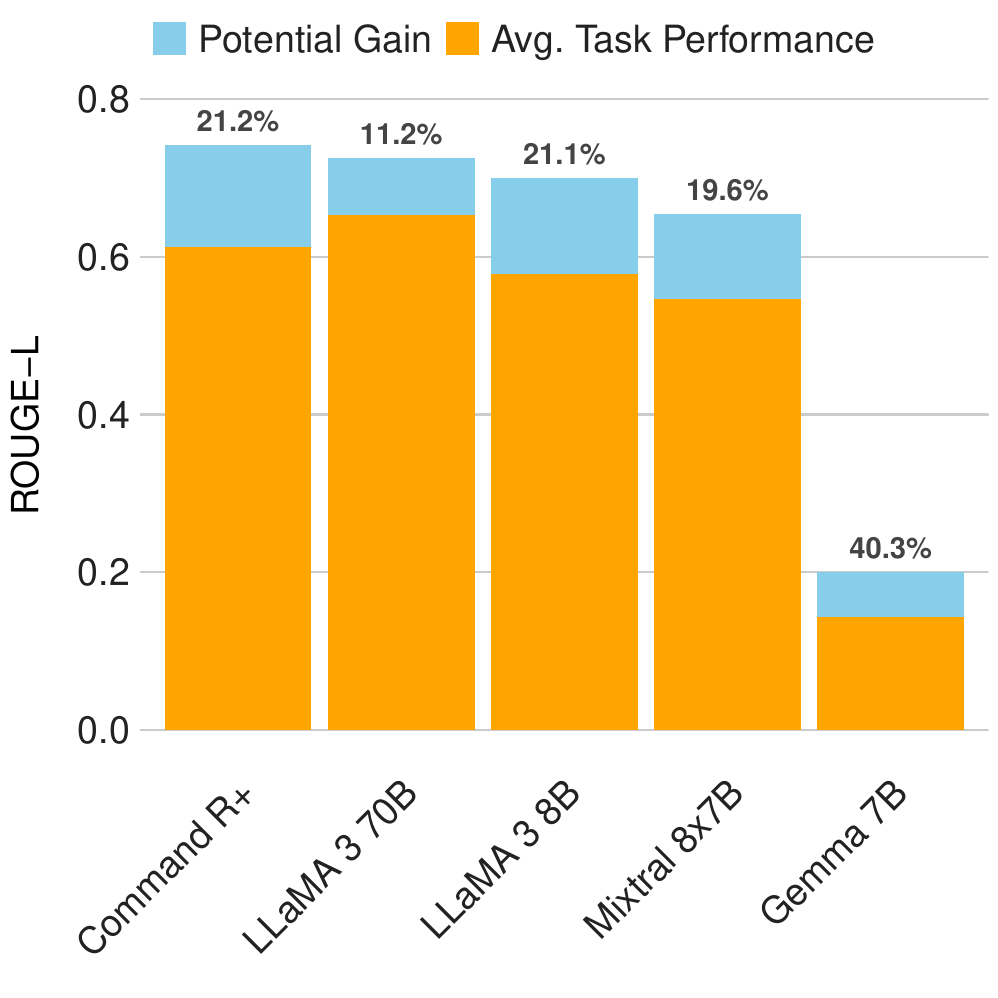}
    \caption{The maximum performance gain models can experience if their prompts are adjusted in optimal ways as explained in \Cref{tab:ap_potential_gain_explained}.}    \label{fig:teaser_max}
\end{figure}

\begin{figure*}[t]
    \begin{AIbox}{Prompt Example for Numerical Commonsense Reasoning}
    \parbox[t]{\linewidth}{
            {\small \texttt{<system>}}\\
            \hspace*{5mm}{\small You are a helpful assistant in solving various tasks. You should only output one answer to the task,\\
            \hspace*{5mm}nothing more, no explanations, and nothing around. Just read the instruction carefully,\\
            \hspace*{5mm}understand the positive and negative examples provided,\\
            \hspace*{5mm}and generate one single answer in the same way as the example's output.}\\
            {\small \texttt{</system>}}
            
            \tcbline

            {\small \texttt{<user>}}\\
            \hspace*{5mm}{\small {\bf Instruction:} Find the most appropriate number to replace the blank (indicated with \textunderscore{}) and express it in words.}\\

            \hspace*{5mm}{\small {\bf Positive examples:}}\\
            {\small
                \hspace*{1cm} Input: A lion has \textunderscore{} legs.\\
                \hspace*{1cm} Output: four\\\\
                \hspace*{1cm} Input: Numbers less than \textunderscore{} are called negative.\\
                \hspace*{1cm} Output: zero\\\\
                \hspace*{1cm} Input: There are \textunderscore{} hours in a day.\\
                \hspace*{1cm} Output: twenty four\\\\
            }

            \hspace*{5mm}{\small {\bf Negative examples:}}\\
            {\small
                \hspace*{1cm} Input: A dog has \textunderscore{} legs.\\
                \hspace*{1cm} Output: 4 \hspace*{1mm} \textcolor{gray}{\# Not expressed in words but numbers.}\\\\
                \hspace*{1cm} Input: Numbers less than \textunderscore{} are called negative.\\
                \hspace*{1cm} Output: one \hspace*{1mm} \textcolor{gray}{\# Logically wrong.}\\
            }
        
            \hspace*{5mm}{\small {\bf Input:} Some plant varieties can grow up to \textunderscore{} feet tall.}\\
            \hspace*{5mm}{\small {\bf Output:}}\\
            {\small\texttt{</user>}}
    }
    \end{AIbox}
    \caption{A prompt example for the first instance in the NumerSense dataset \cite{lin2020birds}.}
    \label{fig:ap_prompt_example_numersense}
\end{figure*}

\begin{table*}[ht]
    \centering
    \resizebox{\textwidth}{!}{
    \begin{tabular}{llc}
        \toprule
        \textbf{Instruction Type} & \textbf{Instruction Text} & \textbf{ROUGE-L} \\
        \midrule
        \multicolumn{3}{l}{\textbf{Example 1: Addition/Deletion}} \\
        Original   & Your task is to generate a headline (title) for this article.  & 0.60 \\
        Paraphrased & Your task is to generate a concise headline (title) for this article. & 0.78 \\
        \midrule
        \multicolumn{3}{l}{\textbf{Example 2: Ellipsis}} \\
        Original   & Your task is to generate a headline (title) for this article.  & 0.60 \\
        Paraphrased & Generate a headline (title) for this article.  & 0.67 \\
        \midrule
        \multicolumn{3}{l}{\textbf{Example 3: Same Polarity Substitution (contextual)}} \\
        Original   & Summarize the main points of the given text in 3-4 sentences. & 0.55 \\
        Paraphrased & Condense the key ideas of the provided passage into 3-4 sentences. & 0.72 \\
        \midrule
        \multicolumn{3}{l}{\textbf{Example 4: Syntax/discourse structure changes}} \\
        Original & Explain the concept of gravity as if you were teaching a 10-year-old child. & 0.63 \\
        Paraphrased & Imagine you're teaching a 10-year-old child. Explain the concept of gravity. & 0.71 \\
        \midrule
        \multicolumn{3}{l}{\textbf{Example 5: Synthetic/analytic substitution}} \\
        Original   & Translate the following English text into French. & 0.58 \\
        Paraphrased & Convert the given English passage to its French equivalent. & 0.70 \\
        \bottomrule
    \end{tabular}
    }
    \caption{Comparison of original and paraphrased instructions for five types with downstream ROUGE-L scores.}
    \label{tab:example_paraphrases}
\end{table*}

\section{Details on the Metrics}
\label{ap:metrics}
We provide the mathematical equations and interpretations of metrics we used to compute various aspects of this study, such as lexical diversity, prompt complexity, potential gains, etc. in the following subsections.

\subsection{Prompt Complexity Metrics}
\noindent\textbf{Absolute Token Deviation.} An intuitive heuristic that measures the total number of tokens added, removed, or changed between two text pairs. This metric reflects the overall extent of alteration in content and can be computed by:

\begin{equation}
    tok(s1, s2) = |N_{s1} - N_{s2}|
\end{equation}

\noindent where $N_s$ represents the number of tokens in sentence $s$. This metric captures the absolute difference in the length of the texts, offering a simple yet effective way to quantify the overall size of changes.

\noindent\textbf{Word Position Deviation.} This metric measures structural alterations by calculating the average shift in the positions of common words between two paraphrased sentences. The equation uses the mean of the maximum relative position shifts of all common words between two sentences $s1$ and $s2$, represented as

\begin{align}
    pos(s1, s2) = & \\ 
    \frac{1}{N_{\mathcal{C}}} \sum & max\{\delta_{s1,s2}(W), \delta_{s2,s1}(W)\}    
\end{align}

\noindent where $\delta_{s1,s2}$ is the relative position shift of a word $W$ with respect to sentence $s1$ in paraphrase pair $(s1, s2)$, and $NC$ is the count of common words.

\noindent\textbf{Lexical Deviation.} This measure quantifies the difference in vocabulary used between two sentences. It is defined as the complement of the ratio of the number of common words to the total number of unique words in both sentences, given by the equation 

\begin{eqnarray}
    lex(s1, s2) = 1 - \frac{N_{\mathcal{C}}}{N_{\mathcal{A}}}
\end{eqnarray}

\noindent where $N_\mathcal{A}$ is the count of all unique words that occur in either occurs in either or both sentences and $N_{\mathcal{C}}$ is the set of common words that occur in both sentences.

\subsection{Prompt Complexity Correlations}
\label{sec:prompt_complexity_calculation}
We calculate the Pearson correlation between each of the above prompt complexity metrics and the downstream performance in the following way.

$P_0$ represents the original prompt and $P_1, P_2, \ldots, P_{26}$ denote the paraphrased versions. 
The performance score for each paraphrased prompt $P_i$ is given by $S_i$, where $S_i$ is ROUGE-L in this study. 
For each paraphrase $P_i$, the marker of change relative to $P_0$ is denoted by $T_i$ ($T_1$ being deviation in absolute number of tokens, $T_2$ word position deviation, and $T_3$ lexical deviation). 
We calculate Pearson correlation $r$ between $T_i$ and task performance $S_j$:

\begin{align}
    Pea&rson(i, j) =  \\ 
    &\frac{\sum_{i=1}^{n} (T_i - \bar{T}) (S_i - \bar{S})}{\sqrt{\sum_{i=1}^{n} (T_i - \bar{T})^2} \sqrt{\sum_{i=1}^{n} (S_i - \bar{S})^2}}
\end{align}

\noindent where $n$ represents the number of paired scores available (26 paraphrase-original pairs), $\bar{T}$ is the mean value of all $n$ markers of change, and $\bar{S}$ is the mean value of all $n$ performance scores.

\subsection{Lexical Diversity Metrics}
We use four main metrics to assess lexical richness for LLM generations: Root Type-Token Ratio (RTTR), Maas, Moving Average Type-Token Ratio (MATTR), and Measure of Lexical Diversity (MTLD). RTTR and Maas metrics consider the entire set of texts for a given category, focusing on the number of distinct words relative to the total number of words, with modifications to account for text length.

\noindent\textbf{RTTR.} Root Type-Token Ratio measures the proportion of unique words in a text, normalized by text length, indicating the lexical variety of the text \cite{guiraud1960rttr}. Computing the quotient of distinct words and the total number of tokens, the metric is defined as:

\[
\text{RTTR} = \frac{T}{\sqrt{N}}
\]

\noindent where \( T \) represents the number of unique types (distinct words), and \( N \) is the total number of tokens (words). 

\noindent\textbf{Maas.} This metric measures the lexical richness by considering the relationship between the number of distinct words and the total word count, using a logarithmic transformation to reduce the impact of text length \cite{maas1972maas}. The measure is calculated using the formula:

\[
\log(V) = \log(N) + \alpha \log(N)
\]

\noindent where \( V \) is the number of distinct words, \( N \) is the total number of words, and \( \alpha \) is a parameter calculated as follows:

\[
\alpha = \frac{\log(V) - \log(N)}{\log(N)}
\]

\noindent\textbf{MATTR.} Moving Average Type-Token Ratio measures the stability of lexical diversity across different text segments, providing an average diversity score that accounts for variations within the text \cite{covington2010cutting}. Taking the average Type-Token Ratio (TTR) over a sliding window of size \( w \), which in our experiments is 25 tokens, is calculated as:

\[
\text{TTR}_i = \frac{T_i}{N_i}
\]

\noindent where \( T_i \) is the number of unique types within the window \( i \), and \( N_i \) is the number of tokens within the window \( i \). The MATTR is then the average of these TTR values over all windows:

\[
\text{MATTR} = \frac{1}{m} \sum_{i=1}^{m} \text{TTR}_i
\]

\noindent where \( m \) is the number of windows. 

\noindent\textbf{MTLD.} Measure of Lexical Diversity determines the text's lexical diversity by determining how many words are needed before the diversity drops below a set threshold, thereby capturing the consistency of word usage variety throughout the text \cite{mccarthy2010mtld}. MTLD measures the text length until the Type-Token Ratio (TTR) reaches a threshold (0.72 in our experiments), then starting a new segment and continuing this process through the entire text measures MTLD. Mathematically:

\[
\text{MTLD} = \frac{\text{Total tokens}}{\text{Number of segments}}
\]

\noindent where a segment is defined as the number of tokens until the TTR reaches the threshold.

Higher RTTR (ranging from 0 to the square root of the total number of tokens), MATTR (0 to 1), and MTLD (typically 0 to infinity) values indicate greater lexical diversity, while lower values suggest less variety in word usage. The Maas metric, which typically ranges close to 0, operates inversely: lower values indicate higher lexical richness, while higher values suggest lower diversity.

\subsection{Downstream Task Metrics}
Because we evaluate on the benchmark suite of \citet{wang-etal-2022-super}, we use the corresponding evaluation metric ROUGE-L for each task. 
Their study shows that ROUGE-L correlates well with human judgments for these tasks and yields comparable scores for all generative tasks. Further, this choice requires no postprocessing or judgment by another model, which could have its own biases.

\subsection{Potential Downstream Gain}
\Cref{fig:teaser} has shown the potential median gain that can be reached by paraphrasing prompts in specific types. In \Cref{tab:ap_potential_gain_explained}, we show how the median is calculated per example. For each example in a task dataset, we take the downstream task scores of paraphrased prompts that are higher than the original prompt and take the median. Then, for all examples in the dataset, we average that median.

Further, one can also compute the maximum performance that can be reached if the best paraphrase type would be chosen to paraphrase the prompt. As previous experiments have shown, it is challenging to devise a robust method to select the best paraphrase of a prompt. More research on the Monte-Carlo approach could bring us closer to this maximum. The median is a fair way to represent a gain that users could expect by selecting one of the multiple paraphrases of prompts that are better than the original. Compared to the mean, the median is not influenced by outliers that are unlikely to select (e.g., the two 1.0 scores in \Cref{tab:ap_potential_gain_explained} are unlikely to reach; the most likely expectation is 0.67, which is also the median).

\begin{table}[t!]
    \centering
    \normalsize
    \begin{tabular}{lrrrH}
    \toprule
    \textbf{Task Family ($\downarrow$)} & \textbf{Tok.} & \textbf{Pos.} & \textbf{Lex.} & \textbf{Avg. p} \\
    \midrule
    Answerability Class. & -0.06 & -0.06 & -0.06 & 0.06 \\
    Commonsense Class. & -0.01 & -0.01 & 0.00 & 0.11 \\
    Coreference Resolution & 0.00 & 0.00 & 0.00 & 0.10 \\
    Dialogue Generation & -0.02 & -0.01 & 0.00 & 0.10 \\
    Fill in The Blank & 0.04 & 0.04 & 0.04 & 0.08 \\
    Information Extr. & 0.03 & 0.06 & 0.04 & 0.16 \\
    Named Entity Rec. & 0.01 & 0.02 & 0.01 & 0.06 \\
    Program Execution & -0.02 & -0.01 & -0.03 & 0.33 \\
    Question Answering & 0.14 & 0.13 & 0.13 & 0.04 \\
    Question Generation & 0.06 & 0.06 & 0.06 & 0.12 \\
    Question Rewriting & 0.03 & 0.04 & 0.04 & 0.28 \\
    Question Underst. & -0.01 & -0.03 & -0.02 & 0.08 \\
    Sentence Composition & 0.10 & 0.11 & 0.11 & 0.04 \\
    Sentiment Analysis & 0.03 & 0.03 & 0.03 & 0.06 \\
    Summarization & 0.08 & 0.08 & 0.08 & 0.01 \\
    Text Categorization & -0.03 & -0.02 & -0.02 & 0.28 \\
    Text Completion & 0.01 & 0.02 & 0.00 & 0.23 \\
    Text Matching & 0.00 & 0.01 & 0.01 & 0.04 \\
    Text Quality Eval. & 0.00 & -0.01 & -0.01 & 0.44 \\
    Text to Code & -0.02 & -0.01 & 0.01 & 0.29 \\
    Textual Entailment & 0.03 & 0.01 & 0.03 & 0.23 \\
    Title Generation & 0.01 & 0.01 & 0.01 & 0.29 \\
    Word Semantics & -0.02 & -0.03 & -0.01 & 0.08 \\
    Wrong Candidate Gen. & 0.01 & 0.03 & 0.00 & 0.15 \\
    Overall & 0.02 & 0.02 & 0.02 & 0.15 \\   
    \bottomrule
    \end{tabular}
\caption{The average Pearson correlations between the number of tokens (Tok.) and task perf., word position deviation (Pos.) and task perf., and lexical deviation (Lex.) and task perf. for LLaMA 3 70B. Each task family has five downstream tasks. The average p-value is 0.05.}
\label{tab:prompt_complexity}
\end{table}

\begin{table}[h!]
\centering
\begin{tabular}{lccc}
\toprule
\textbf{Task Family} & \textbf{Min} & \textbf{Max} & \textbf{Avg ($\downarrow$)} \\
\midrule
Title Generation & 24  & 53   & 36.0  \\
Fill in The Blank & 25  & 60   & 43.2  \\
Summarization & 21  & 123  & 52.0  \\
Text Quality Eval. & 42  & 67   & 52.6  \\
Named Entity Rec. & 38  & 77   & 52.8  \\
Word Semantics & 39  & 74   & 53.8  \\
Sentiment Analysis & 24  & 115  & 56.6  \\
Sentence Composition & 19  & 84   & 61.8  \\
Answerability Class. & 47  & 70   & 63.6  \\
Text Completion & 40  & 111  & 66.0  \\
Dialogue Generation & 27  & 186  & 67.8  \\
Question Answering & 45  & 106  & 72.0  \\
Program Execution & 43  & 110  & 75.4  \\
Coreference Resolution & 24  & 219  & 78.8  \\
Information Extr. & 70  & 91   & 79.0  \\
Textual Entailment & 61  & 111  & 81.2  \\
Text Categorization & 55  & 116  & 88.2  \\
Text Matching & 34  & 203  & 91.0  \\
Question Generation & 29  & 160  & 102.2 \\
Wrong Candidate Gen. & 105 & 192  & 129.4 \\
Commonsense Class. & 49  & 202  & 148.8 \\
Question Rewriting & 46  & 753  & 216.0 \\
Question Understanding & 125 & 581  & 291.8 \\
Text to Code & 121 & 638  & 390.6 \\
Overall & 19  & 753  & 102.1 \\
\bottomrule
\end{tabular}
\caption{The number of tokens per input prompt (computed using the GPT-2 tokenizer). \textbf{Min} shows the shortest prompt, \textbf{Max} the longest prompt, and \textbf{Avg} the average prompt length for all five tasks.}
\label{tab:tokens_per_task}
\end{table}

\section{Additional Results}
\label{ap:full_results}
\noindent\textbf{Prompt Complexity Correlations.} \Cref{tab:prompt_complexity} shows the individual correlations between the three markers of prompt complexity and downstream task performance as described in \Cref{sec:prompt_complexity_calculation}. All correlations are close to zero, and only very small correlations exist for question answering (0.13 - 0.14) and sentence composition (0.10 - 0.11) with $p<0.05$. We could generally find no evidence that the complexity of the prompt contributes to gains or losses in performance outcomes.

\noindent\textbf{Prompt Lengths.} We conducted a post-hoc analysis of the number of input tokens across various task families using the GPT-2 tokenizer. As shown in \Cref{tab:tokens_per_task}, the average number of tokens varies significantly depending on the nature of the task, ranging from 36 tokens for title generation tasks to 390 tokens for text-to-code tasks. The overall distribution of input tokens is even broader when looking at individual tasks, with the shortest task, sentence composition, using as few as 19 tokens, and the longest, question rewriting, reaching up to 753 tokens. This wide spectrum of prompt lengths indicates that our study covers a diverse array of real-world use cases, from concise to extended prompts. By analyzing 120 tasks across 24 task families, we believe our conclusions hold across a broad range of prompt lengths, including those much shorter (e.g., fewer than 36 tokens) and much longer (e.g., more than 800 tokens).

\noindent\textbf{Results by Task and Type.} We show a detailed table of avgerage downstream task performance gain or loss over the original prompt of a task when paraphrasing the prompt using modifications of one of the six different paraphrase groups in \Cref{tab:ap_task_by_types}. 
We decompose this table for each model in \Cref{tab:ap_task_by_types_gemma,tab:ap_task_by_types_mixtral,tab:ap_task_by_types_command_r_plus,tab:ap_task_by_types_llama_70b,tab:ap_task_by_types_llama_8b}.

\noindent\textbf{Maximum Potential Gain.} To understand how much models would possibly benefit from adjusting prompts in optimal ways as explained in \Cref{tab:ap_potential_gain_explained}, we also show the maximum potential gain in \Cref{fig:teaser_max} of the teaser on the first page of this paper (\Cref{fig:teaser}). 
This gives an upper bound to what is possible if prompts were optimally adjusted with paraphrase types. 
Again, we note that although finding these optimal prompt adjustments is challenging, this task is not impossible and gives space for future work.

\begin{figure*}
    \centering
    \includegraphics[width=\textwidth]{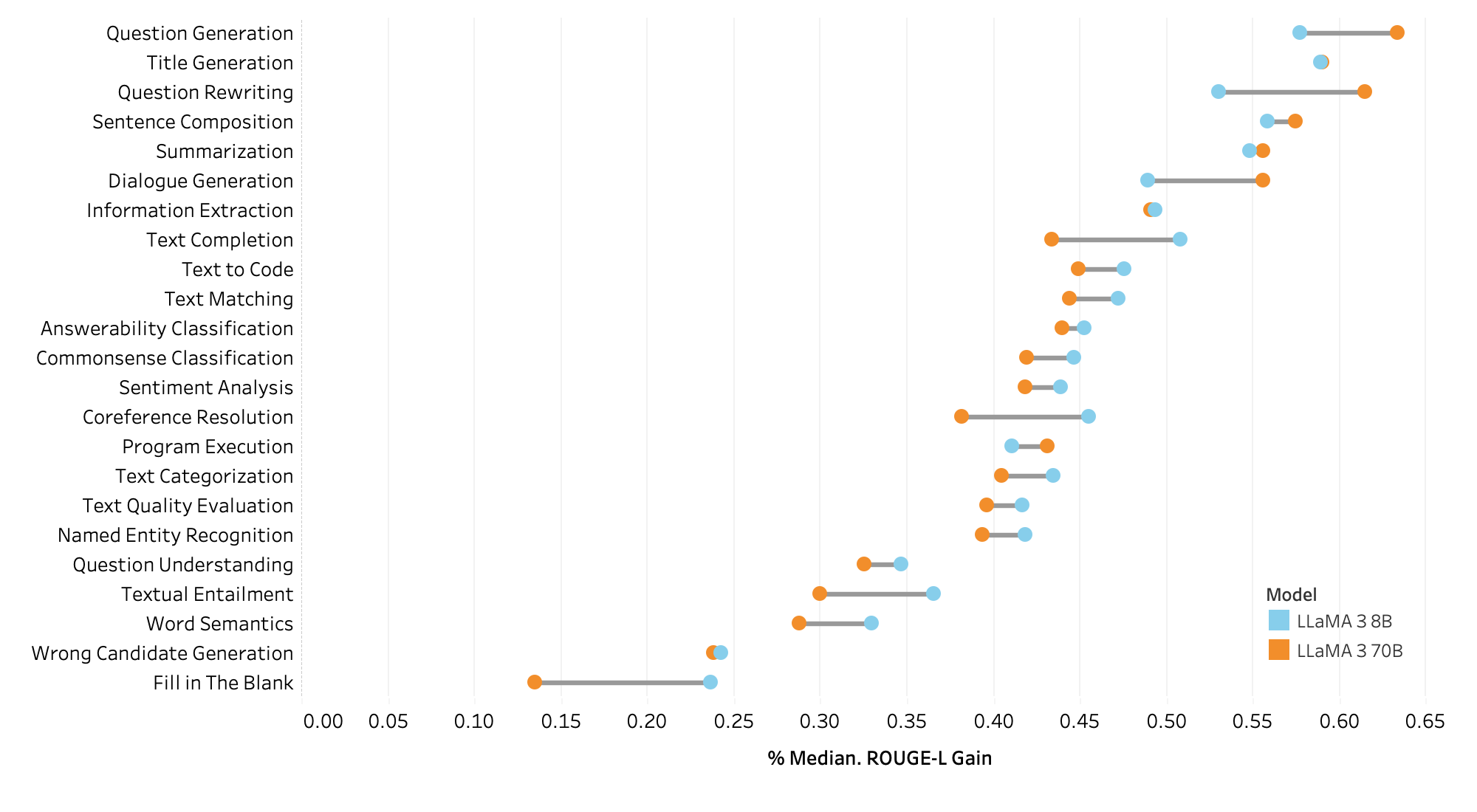}
    \caption{A comparison between model scale for LLaMA 3 8B (blue) and 70B (orange) for different tasks.}
    \label{fig:q3_llama_scale_comparison}
\end{figure*}

\noindent\textbf{Model Scale Comparison.} \Cref{fig:q3_llama_scale_comparison} compares model size between LLaMA 3 8B and LLaMA 3 70B. 
Observe how the smaller 8B parameter model benefits more across tasks when compared to its larger 70B version. 
Only for program execution, information extraction, sentence completion, question rewriting, and question generation, the 70B model shows a larger gain.

Upon closer examination, the performance difference between the two models across various tasks highlights several trends. The smaller 8B model consistently outperforms the 70B model in tasks such as text categorization, text quality evaluation, named entity recognition, and sentiment analysis. This suggests that for these specific tasks, the architectural or training enhancements in the 8B model may be more efficiently leveraged, potentially due to better optimization or more effective use of parameter space.

In contrast, the 70B model demonstrates improvements over the 8B model in tasks like program execution, information extraction, and sentence completion. This indicates that for these more complex or nuanced tasks, the increased parameter count of the 70B model likely provides a richer representation space, allowing it to capture and utilize more intricate patterns and dependencies in the data.

\begin{figure*}
    \centering
    \includegraphics[width=\textwidth]{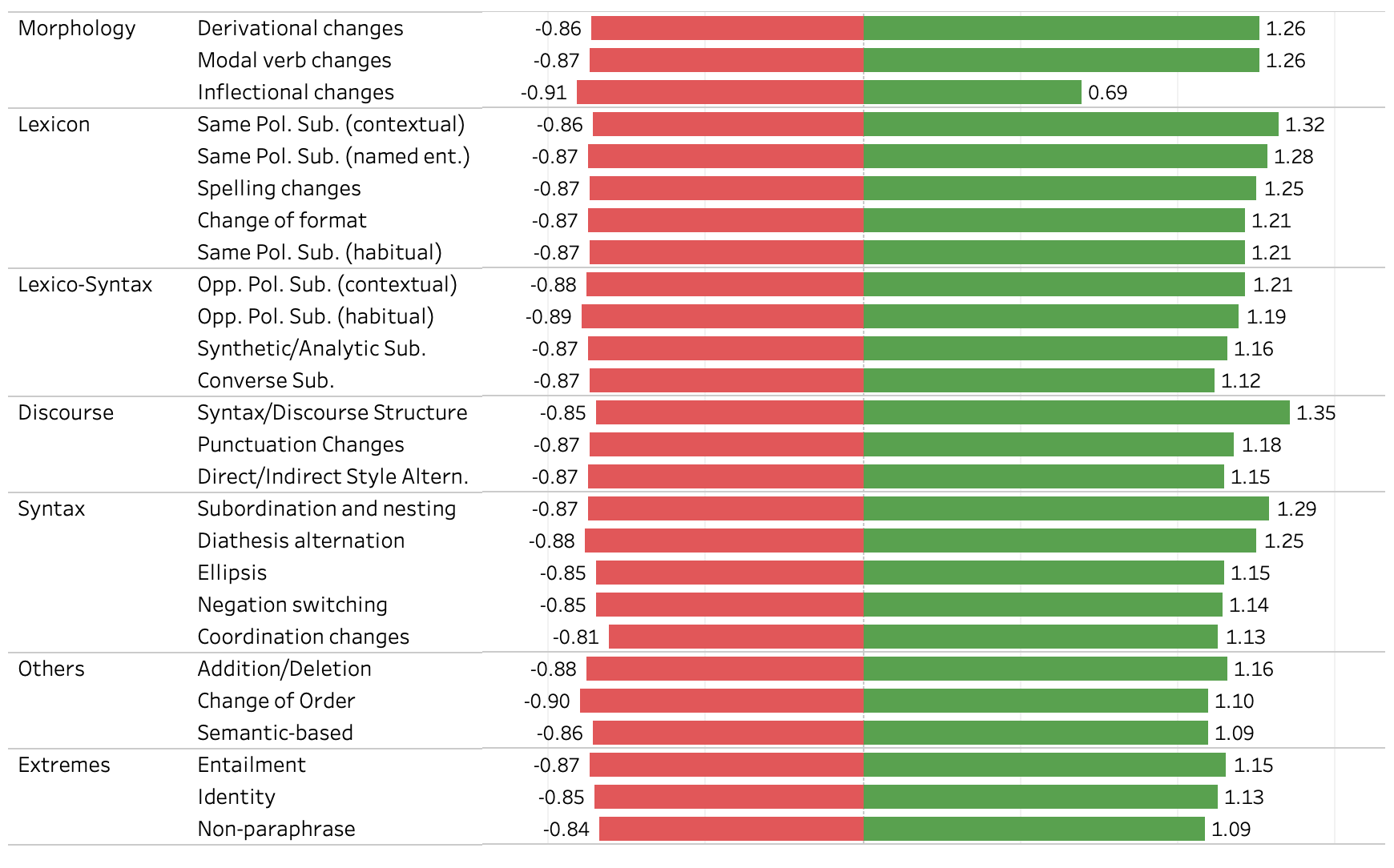}
    \caption{The average percentage in downstream task performance gain or loss for applying different paraphrase types across all five models. This decomposes the groups of \Cref{fig:q1_dumbbell} into their individual types.}
    \label{fig:ap_gain_loss_groups_types}
\end{figure*}

\begin{table*}[t]
\centering
\begin{tabular}{@{}llHHlrrlH@{}}
\toprule
\textbf{Task ID} & \textbf{Dataset Name} & \textbf{Sources} & \textbf{Task Categories} & \textbf{Domain} & \textbf{Inp. Tokens} & \textbf{Out. Tokens} & \textbf{URL} & \textbf{Reasoning Categories} \\ \midrule
\multicolumn{9}{l}{\textbf{Text Completion} - Morphology (\textcolor{forestgreen}{+11.4\% $\uparrow$}), Syntax \textcolor{red}{(-5.3\% $\downarrow$})}\\
\quad 105 & ROCStories & pib & Text Completion & Story, Commonsense & 39.3 & 9.0 & \href{https://huggingface.co/datasets/pib}{Link} & Commonsense Reasoning \\
\quad 297 & ROCStories & roc\_stories & Text Completion & Story & 56.1 & 1.0 & \href{https://cs.rochester.edu/nlp/rocstories/}{Link} &  \\
\quad 268 & CaseHold & casehold & Text Completion & Law & 269.1 & 1.0 & \href{https://github.com/reglab/casehold}{Link} & Logical Reasoning \\
\quad 138 & Detoxifying LMs & detoxifying\_lms & Text Completion & Social Media & 59.6 & 2.0 & \href{https://aclanthology.org/2021.naacl-main.190/}{Link} &  \\
\quad 296 & ROCStories & roc\_stories & Text Completion & Story & 56.1 & 1.0 & \href{https://cs.rochester.edu/nlp/rocstories/}{Link} &  \\
\midrule
\multicolumn{9}{l}{\textbf{Question Answering} - Extremes (\textcolor{forestgreen}{+24.8\% $\uparrow$}), Lexicon (\textcolor{red}{-3.0\% $\downarrow$}) } \\
\quad 1441 & DoQA Movies & doqa & Question Answering & Movies, Dialogue & 190.6 & 16.0 & \href{http://www.ixa.eus/node/12931}{Link} & Reasoning on Social Interactions \\
\quad 2 & Quoref & quoref & Question Answering & Wikipedia & 356.6 & 1.7 & \href{https://allenai.org/data/quoref}{Link} &  \\
\quad 24 & Cosmos QA & cosmosqa & Question Answering & Personal Narratives & 82.5 & 8.3 & \href{https://leaderboard.allenai.org/cosmosqa/submissions/about}{Link} & Commonsense Reasoning \\
\quad 332 & TellMeWhy & tellmewhy & Question Answering & Story & 50.4 & 8.9 & \href{https://github.com/StonyBrookNLP/tellmewhy}{Link} & Abductive Reasoning \\
\quad 310 & RACE & race & Question Answering & English Exams & 351.6 & 1.0 & \href{http://www.cs.cmu.edu/~glai1/data/race/}{Link} & Scientific Reasoning \\
\midrule
\multicolumn{9}{l}{\textbf{Fill in The Blank} - Extremes (\textcolor{forestgreen}{+18.4\% $\uparrow$}), Discourse (\textcolor{red}{-5.0\% $\downarrow$})}\\
\quad 672 & NumerSense & numersense & Fill in The Blank & Commonsense & 11.1 & 1.0 & \href{https://inklab.usc.edu/NumerSense/}{Link} & Commonsense Reasoning -> Numerical Commonsense Reasoning \\
\quad 277 & StereoSet & stereoset & Fill in The Blank & Stereotypes & 8.8 & 1.0 & \href{https://stereoset.mit.edu}{Link} & Commonsense Reasoning \\
\quad 965 & LibriSpeech ASR & librispeech\_asr & Fill in The Blank & Books & 42.1 & 1.0 & \href{https://huggingface.co/datasets/librispeech\_asr}{Link} & Abductive Reasoning \\
\quad 1217 & ATOMIC & atomic & Fill in The Blank & Sociology, Commonsense & 5.3 & 1.1 & \href{https://allenai.org/data/atomic-2020}{Link} & Relational Reasoning, Reasoning on Social Interactions, Commonsense Reasoning \\
\quad 1360 & NumerSense & numersense & Fill in The Blank & Concepts and Relations & 26.1 & 1.0 & \href{https://inklab.usc.edu/NumerSense/}{Link} & Commonsense Reasoning \\
\midrule
\multicolumn{9}{l}{\textbf{Sentiment Analysis} - Morphology (\textcolor{forestgreen}{+15.6\% $\uparrow$}), Discourse (\textcolor{red}{-2.1\% $\downarrow$})} \\
\quad 420 & PerSenT & persent & Sentiment Analysis & News & 352.6 & 1.0 & \href{https://github.com/MHDBST/PerSenT}{Link} &  \\
\quad 889 & GoEmotions & go\_emotions & Sentiment Analysis & Narrative, Dialogue & 12.3 & 1.0 & \href{https://huggingface.co/datasets/go\_emotions}{Link} & Reasoning on Social Interactions \\
\quad 1312 & Amazon Review Polarity & multilingual\_amazon\_reviews & Sentiment Analysis & Reviews & 56.7 & 1.0 & \href{https://huggingface.co/datasets/amazon\_reviews\_multi}{Link} & Reasoning on Social Interactions, Commonsense Reasoning \\
\quad 1535 & Daily Dialog & dailydialog & Sentiment Analysis & Dialogue & 137.8 & 1.0 & \href{https://huggingface.co/datasets/daily\_dialog}{Link} &  \\
\quad 284 & IMDB & imdb & Sentiment Analysis & Movie Reviews & 229.9 & 1.0 & \href{https://ai.stanford.edu/~amaas/data/sentiment/}{Link} &  \\
\midrule
\multicolumn{9}{l}{\textbf{Program Execution} - Morphology (\textcolor{forestgreen}{+1.9\% $\uparrow$}), Discourse (\textcolor{red}{-4.2\% $\downarrow$})} \\
\quad 208 & Combinations of List & synthetic & Program Execution & Mathematics & 5.0 & 22.1 & - & Numerical Reasoning, Mathematics -> Combinatorics \\
\quad 378 & Reverse Words & synthetic & Program Execution & Image Captions & 21.4 & 10.4 & - & Mathematics -> Counting \\
\quad 1148 & Maximum ASCII Value & synthetic & Program Execution & Computer Science & 1.0 & 1.0 & - &  \\
\quad 94 & Calculate Mean & conala & Program Execution & Code & 6.0 & 1.0 & \href{https://arxiv.org/pdf/1805.08949.pdf}{Link} & Quantitative Reasoning, Numerical Reasoning, Mathematics -> Arithmetic, Mathematics -> Statistics \\
\quad 99 & Reverse Elements & synthetic & Program Execution & Code & 17.9 & 6.3 & - &  \\
\midrule
\multicolumn{9}{l}{\textbf{Question Generation} - Morphology (\textcolor{forestgreen}{+21.5\% $\uparrow$}), Discourse (\textcolor{forestgreen}{+1.3\% $\uparrow$})} \\
\quad 6 & MCTACO & mctaco & Question Generation & News, Wiki, Law, History & 18.9 & 8.2 & \href{https://github.com/CogComp/MCTACO}{Link} & Temporal Reasoning, Commonsense Reasoning \\
\quad 1602 & WebQuestions & web\_questions & Question Generation & Knowledge Base & 9.6 & 6.6 & \href{https://huggingface.co/datasets/web\_questions}{Link} &  \\
\quad 599 & CUAD & cuad & Question Generation & Law & 3153.9 & 43.3 & \href{https://huggingface.co/datasets/cuad}{Link} &  \\
\quad 405 & NarrativeQA & narrativeqa & Question Generation & Books, Movies & 575.7 & 8.6 & \href{https://github.com/deepmind/narrativeqa}{Link} &  \\
\quad 739 & LhoestQ & - & Question Generation & Web & 142.1 & 10.4 & - & Abductive Reasoning \\
\bottomrule
\end{tabular}
\caption{Overview of the tasks in this study with their average input and output tokens across all examples. We also show the two paraphrase types with the highest and lowest gain or loss across all five sampled tasks within a task category. Table 1/4.}\label{tab:ap_overview_tasks_1}
\end{table*}

\begin{table*}[t]
\centering
\begin{tabular}{@{}llHHlrrlH@{}}
\toprule
\textbf{Task ID} & \textbf{Dataset Name} & \textbf{Sources} & \textbf{Task Categories} & \textbf{Domain} & \textbf{Inp. Tokens} & \textbf{Out. Tokens} & \textbf{URL} & \textbf{Reasoning Categories} \\ \midrule
\multicolumn{9}{l}{\textbf{Text to Code} - Morphology (\textcolor{forestgreen}{+2.4\% $\uparrow$}), Lexico-Syntax \textcolor{red}{(-5.5\% $\downarrow$})}\\
\quad 869 & CFQ MCD1 & cfq\_mcd1 & Text to Code & SQL & 41.2 & 1.0 & \href{https://www.tensorflow.org/datasets/catalog/cfq}{Link} &  \\
\quad 956 & LeetCode 420 & leetcode & Text to Code & Mathematics & 3.0 & 1.0 & \href{https://leetcode.com/problems/strong-password-checker/}{Link} & Logical Reasoning \\
\quad 128 & SCAN & scan & Text to Code & Machine Learning & 7.0 & 10.8 & \href{https://github.com/brendenlake/SCAN}{Link} &  \\
\quad 107 & SPLASH & splash & Text to Code & SQL & 13.3 & 17.5 & \href{https://arxiv.org/pdf/2005.02539.pdf}{Link} &  \\
\quad 211 & Logic2Text & logic2text & Text to Code & Wikipedia, Logic & 52.6 & 1.0 & \href{https://github.com/czyssrs/Logic2Text}{Link} & Logical Reasoning \\
\midrule
\multicolumn{9}{l}{\textbf{Question Rewriting} - Morphology (\textcolor{red}{-4.7\% $\downarrow$}), Syntax \textcolor{red}{(-17.2\% $\downarrow$})}\\
\quad 402 & GrailQA & grailqa & Question Rewriting & Knowledge Base & 125.5 & 11.2 & \href{https://dki-lab.github.io/GrailQA/}{Link} &  \\
\quad 671 & AmbigQA & ambigqa & Question Rewriting & Wikipedia & 8.9 & 14.4 & \href{https://nlp.cs.washington.edu/ambigqa/}{Link} & Abductive Reasoning \\
\quad 34 & WinoGrande & winogrande & Question Rewriting & Commonsense & 26.1 & 20.0 & \href{https://leaderboard.allenai.org/winogrande/submissions/about}{Link} & Commonsense Reasoning \\
\quad 1622 & Disfl QA & disfl\_qa & Question Rewriting & Wikipedia & 14.6 & 9.9 & \href{https://huggingface.co/datasets/disfl\_qa}{Link} &  \\
\quad 670 & AmbigQA & ambigqa & Question Rewriting & Wikipedia & 8.9 & 12.1 & \href{https://nlp.cs.washington.edu/ambigqa/}{Link} & Abductive Reasoning \\
\midrule
\multicolumn{9}{l}{\textbf{Summarization} - Lexicon (\textcolor{forestgreen}{+10.8\% $\uparrow$}), Syntax \textcolor{forestgreen}{(+2.6\% $\uparrow$})}\\
\quad 1290 & XSum & xsum & Summarization & News & 371.8 & 21.1 & \href{https://github.com/EdinburghNLP/XSum/tree/master/XSum-Dataset}{Link} &  \\
\quad 1357 & XLSum & xlsum & Summarization & News & 459.2 & 22.1 & \href{https://arxiv.org/abs/2106.13822}{Link} &  \\
\quad 668 & SciTLDR & scitldr & Summarization & Scientific Papers & 159.5 & 20.6 & \href{https://github.com/allenai/scitldr}{Link} &  \\
\quad 589 & Amazon Food Reviews & amazon\_fine\_food\_reviews & Summarization & Reviews & 79.6 & 4.2 & \href{https://metatext.io/datasets/amazon-fine-food-reviews}{Link} &  \\
\quad 672 & Amazon / Yelp Summ. & amazon\_and\_yelp\_summarization\_dataset & Summarization & Reviews & 404.0 & 50.7 & \href{https://github.com/abrazinskas/FewSum}{Link} & Reasoning on Social Interactions \\
\midrule
\multicolumn{9}{l}{\textbf{Commonsense Classification} - Morphology (\textcolor{forestgreen}{+14.6\% $\uparrow$}), Lexicon \textcolor{red}{(-2.7\% $\downarrow$})}\\
\quad 116 & Com2Sense & com2sense & Commonsense Classification & Concepts and Relations & 19.2 & 1.0 & \href{https://github.com/PlusLabNLP/Com2Sense}{Link} & Commonsense Reasoning \\
\quad 1204 & ATOMIC & atomic & Commonsense Classification & Social Commonsense & 10.4 & 1.0 & \href{https://allenai.org/data/atomic-2020}{Link} & Relational Reasoning, Reasoning on Social Interactions, Commonsense Reasoning -> Social Situations \\
\quad 1209 & ATOMIC & atomic & Commonsense Classification & Physical Commonsense & 7.0 & 1.0 & \href{https://allenai.org/data/atomic-2020}{Link} & Relational Reasoning, Reasoning on Social Interactions, Commonsense Reasoning -> Social Situations, Reasoning on Objects \\
\quad 1199 & ATOMIC & atomic & Commonsense Classification & Social Commonsense & 7.4 & 1.0 & \href{https://allenai.org/data/atomic-2020}{Link} & Relational Reasoning, Reasoning on Social Interactions, Commonsense Reasoning -> Social Situations \\
\quad 1208 & ATOMIC & atomic & Commonsense Classification & Social Commonsense & 7.0 & 1.0 & \href{https://allenai.org/data/atomic-2020}{Link} & Relational Reasoning, Reasoning on Social Interactions, Commonsense Reasoning \\
\midrule
\multicolumn{9}{l}{\textbf{Text Matching} - Syntax (\textcolor{forestgreen}{+5.4\% $\uparrow$}), Discourse \textcolor{red}{(-12.2\% $\downarrow$})}\\
\quad 1288 & GLUE MRPC & mrpc & Text Matching & News, Web & 45.9 & 1.0 & \href{https://www.microsoft.com/en-us/download/details.aspx?id=52398}{Link} & Analogical Reasoning \\
\quad 276 & Enhanced WSC & wsc; enhanced\_wsc & Text Matching & Dialogue, Narrative & 39.8 & 1.0 & \href{https://huggingface.co/datasets/winograd\_wsc; https://github.com/mhany90/perturbed-wsc}{Link} & Commonsense Reasoning \\
\quad 910 & Bianet & bianet & Text Matching & News & 48.5 & 1.0 & \href{https://huggingface.co/datasets/bianet}{Link} &  \\
\quad 148 & AFS & afs & Text Matching & Government and Politics & 25.6 & 1.0 & \href{https://arxiv.org/pdf/1709.01887.pdf}{Link} &  \\
\quad 624 & OHSUMED & ohsumed & Text Matching & Scientific Papers & 188.5 & 11.2 & \href{https://huggingface.co/datasets/ohsumed}{Link} &  \\
\midrule
\multicolumn{9}{l}{\textbf{Word Semantics} - Morphology (\textcolor{forestgreen}{+14.9\% $\uparrow$}), Lexico-Syntax \textcolor{red}{(-6.5\% $\downarrow$})}\\
\quad 1585 & ROOT09 & root09 & Word Semantics & Misc. & 1.0 & 1.0 & \href{https://aclanthology.org/W16-5304/}{Link} & Reasoning on Objects \\
\quad 1582 & BLESS & bless & Word Semantics & Misc. & 1.0 & 1.0 & \href{https://aclanthology.org/W11-2501/}{Link} & Reasoning on Objects \\
\quad 141 & Odd-Man-Out & odd\_man\_out & Word Semantics & Card Game & 9.0 & 1.1 & \href{https://github.com/gabrielStanovsky/odd-man-out}{Link} &  \\
\quad 142 & Odd-Man-Out & odd\_man\_out & Word Semantics & Card Game & 5.3 & 1.1 & \href{https://github.com/gabrielStanovsky/odd-man-out}{Link} &  \\
\quad 458 & MATRES & matres & Word Semantics & News & 49.2 & 1.0 & \href{https://github.com/CogComp/MATRES}{Link} &  \\
\bottomrule
\end{tabular}
\caption{Overview of the tasks in this study with their average input and output tokens across all examples. We also show the two paraphrase types with the highest and lowest gain or loss across all five sampled tasks within a task category. Table 2/4.}\label{tab:ap_overview_tasks_2}
\end{table*}

\begin{table*}[t]
\centering
\begin{tabular}{@{}llHHlrrlH@{}}
\toprule
\textbf{Task ID} & \textbf{Dataset Name} & \textbf{Sources} & \textbf{Task Categories} & \textbf{Domain} & \textbf{Inp. Tokens} & \textbf{Out. Tokens} & \textbf{URL} & \textbf{Reasoning Categories} \\ \midrule
\multicolumn{9}{l}{\textbf{Question Understanding} - Morphology (\textcolor{forestgreen}{+15.1\% $\uparrow$}), Discourse \textcolor{red}{(-4.1\% $\uparrow$})}\\
\quad 19 & MCTACO & mctaco & Question Understanding & News, Wiki, Law, History & 27.8 & 1.0 & \href{https://github.com/CogComp/MCTACO}{Link} & Temporal Reasoning, Commonsense Reasoning \\
\quad 46 & Misc. & Misc. & Question Understanding & Pop, Nat. Science, History & 14.8 & 1.0 & - & Commonsense Reasoning \\
\quad 18 & MCTACO & mctaco & Question Understanding & News, Wiki, Law, History & 24.9 & 1.0 & \href{https://github.com/CogComp/MCTACO}{Link} & Temporal Reasoning \\
\quad 1289 & TREC & trec & Question Understanding & Misc. & 10.2 & 1.0 & \href{https://cogcomp.seas.upenn.edu/Data/QA/QC/}{Link} &  \\
\midrule
\multicolumn{9}{l}{\textbf{Text Quality Evaluation} - Morphology (\textcolor{forestgreen}{+7.7\% $\uparrow$}), Discourse \textcolor{red}{(-8.6\% $\uparrow$})}\\
\quad 616 & CoLA & cola & Text Quality Evaluation & Linguistics & 7.9 & 1.0 & \href{https://nyu-mll.github.io/CoLA/}{Link} & Grammatical Reasoning \\
\quad 1284 & Human Ratings of NLG & human\_ratings\_of\_natural\_language\_generation\_outputs & Text Quality Evaluation & Dialogue, Restaurants & 23.8 & 1.0 & \href{https://researchportal.hw.ac.uk/en/datasets/human-ratings-of-natural-language-generation-outputs}{Link} & Textual Entailment -> Deductive Reasoning \\
\quad 1623 & Disfl QA & disfl\_qa & Text Quality Evaluation & Wikipedia & 12.2 & 1.0 & \href{https://huggingface.co/datasets/disfl\_qa}{Link} &  \\
\quad 1283 & Human Ratings of NLG & human\_ratings\_of\_natural\_language\_generation\_outputs & Text Quality Evaluation & Dialogue, Restaurants & 24.9 & 1.0 & \href{https://researchportal.hw.ac.uk/en/datasets/human-ratings-of-natural-language-generation-outputs}{Link} &  \\
\quad 1341 & MSR Text Compression & msr\_text\_compression & Text Quality Evaluation & News, Dialogue, Misc. & 19.4 & 1.0 & \href{https://huggingface.co/datasets/msr\_text\_compression}{Link} &  \\
\midrule
\multicolumn{9}{l}{\textbf{Dialogue Generation} - Morphology (\textcolor{forestgreen}{+9.3\% $\uparrow$}), Discourse \textcolor{red}{(-3.9\% $\uparrow$})}\\
\quad 639 & MultiWOZ v2.2 & multi\_woz\_v22 & Dialogue Generation & Dialogue & 13.2 & 12.6 & \href{https://huggingface.co/datasets/multi\_woz\_v22}{Link} & Reasoning on Social Interactions \\
\quad 1730 & PersonaChat & personachat & Dialogue Generation & Dialogue & 143.0 & 9.9 & \href{https://huggingface.co/datasets/bavard/personachat\_truecased}{Link} & Reasoning on Social Interactions, Social Situations \\
\quad 576 & Curiosity Dialogs & curiosity\_dialogs & Dialogue Generation & Concepts and Relations & 66.2 & 23.0 & \href{https://github.com/facebookresearch/curiosity}{Link} & Reasoning on Social Interactions \\
\quad 361 & Spolin & spolin & Dialogue Generation & Dialogue & 32.3 & 1.0 & \href{https://justin-cho.com/spolin}{Link} &  \\
\quad 1603 & SMCalFlow & smcalflow & Dialogue Generation & Dialogue & 8.0 & 8.7 & \href{https://www.mitpressjournals.org/doi/pdf/10.1162/tacl\_a\_00333}{Link} & Social Situations \\
\midrule
\multicolumn{9}{l}{\textbf{Coreference Resolution} - Others (\textcolor{forestgreen}{+25.1\% $\uparrow$}), Lexico-Syntax \textcolor{forestgreen}{(+1.6\% $\uparrow$})}\\
\quad 1391 & WinoGrande & winogrande & Coreference Resolution & Physical Commonsense & 22.8 & 1.0 & \href{https://huggingface.co/datasets/winogrande/}{Link} & Commonsense Reasoning \\
\quad 648 & Winograd WSC & winograd\_wsc & Coreference Resolution & Narrative & 19.4 & 1.7 & \href{https://huggingface.co/datasets/winograd\_wsc}{Link} &  \\
\quad 133 & WinoWhy & winowhy & Coreference Resolution & Concepts and Relations & 43.8 & 1.0 & \href{https://github.com/HKUST-KnowComp/WinoWhy}{Link} & Commonsense Reasoning \\
\quad 330 & GAP & gap & Coreference Resolution & Wikipedia & 73.8 & 1.4 & \href{https://github.com/google-research-datasets/gap-coreference}{Link} &  \\
\quad 329 & GAP & gap & Coreference Resolution & Wikipedia & 81.2 & 1.0 & \href{https://github.com/google-research-datasets/gap-coreference}{Link} &  \\
\midrule
\multicolumn{9}{l}{\textbf{Answerability Classification} - Lexico-Syntax (\textcolor{forestgreen}{+11.8\% $\uparrow$}), Others \textcolor{red}{(-3.5\% $\uparrow$})}\\
\quad 290 & TellMeWhy & tellmewhy & Answerability Classification & Story & 51.0 & 1.3 & \href{https://github.com/StonyBrookNLP/tellmewhy}{Link} &  \\
\quad 50 & MultiRC & multirc & Answerability Classification & News, Wiki, Law, History & 23.3 & 1.0 & \href{https://github.com/CogComp/multirc}{Link} &  \\
\quad 349 & SQuAD 2.0 & squad2.0 & Answerability Classification & Wikipedia & 140.5 & 1.0 & \href{https://arxiv.org/pdf/1806.03822.pdf}{Link} &  \\
\quad 20 & MCTACO & mctaco & Answerability Classification & News, Wiki, Law, History & 25.0 & 1.0 & \href{https://github.com/CogComp/MCTACO}{Link} & Temporal Reasoning, Commonsense Reasoning \\
\quad 1640 & Adversarial QA & adversarial\_qa & Answerability Classification & Wikipedia & 127.3 & 1.0 & \href{https://huggingface.co/datasets/adversarial\_qa}{Link} &  \\
\midrule
\multicolumn{9}{l}{\textbf{Wrong Candidate Generation} - Morphology (\textcolor{forestgreen}{+26.0\% $\uparrow$}), Syntax \textcolor{red}{(-1.4\% $\uparrow$})}\\
\quad 135 & Winowhy & winowhy & Wrong Candidate Generation & Concepts and Relations & 26.0 & 12.5 & \href{https://github.com/HKUST-KnowComp/WinoWhy}{Link} & Commonsense Reasoning \\
\quad 42 & QASC & qasc & Wrong Candidate Generation & Nat. Science & 22.5 & 1.6 & \href{https://allenai.org/data/qasc}{Link} &  \\
\quad 55 & MultiRC & multirc & Wrong Candidate Generation & News, Wiki, Law, History & 308.0 & 3.4 & \href{https://github.com/CogComp/multirc}{Link} & Scientific Reasoning, Multihop Reasoning \\
\quad 11 & MCTACO & mctaco & Wrong Candidate Generation & News, Wiki, Law, History & 27.6 & 4.9 & \href{https://github.com/CogComp/MCTACO}{Link} & Temporal Reasoning, Commonsense Reasoning \\
\quad 631 & DBPedia 14 & dbpedia\_14 & Wrong Candidate Generation & Wikipedia & 53.4 & 1.4 & \href{https://huggingface.co/datasets/dbpedia\_14}{Link} & \\
\bottomrule
\end{tabular}
\caption{Overview of the tasks in this study with their average input and output tokens across all examples. We also show the two paraphrase types with the highest and lowest gain or loss across all five sampled tasks within a task category. Table 3/4.}\label{tab:ap_overview_tasks_3}
\end{table*}

\begin{table*}[t]
\centering
\begin{tabular}{@{}llHHlrrlH@{}}
\toprule
\textbf{Task ID} & \textbf{Dataset Name} & \textbf{Sources} & \textbf{Task Categories} & \textbf{Domain} & \textbf{Inp. Tokens} & \textbf{Out. Tokens} & \textbf{URL} & \textbf{Reasoning Categories} \\ \midrule
\multicolumn{9}{l}{\textbf{Sentence Composition} - Discourse (\textcolor{forestgreen}{+3.8\% $\uparrow$}), Morphology \textcolor{red}{(-5.0\% $\downarrow$})}\\
\quad 184 & SNLI & snli & Sentence Composition & Image Captions & 23.4 & 8.3 & \href{https://nlp.stanford.edu/pubs/snli_paper.pdf}{Link} & Textual Entailment \\
\quad 1613 & SICK & sick & Sentence Composition & Image \& Video Captions & 12.6 & 9.2 & \href{https://huggingface.co/datasets/sick}{Link} & Textual Entailment -> Abductive Reasoning \\
\quad 1530 & SciTailv1.1 & scitailv1.1 & Sentence Composition & Nat. Science & 19.6 & 12.2 & \href{http://data.allenai.org/scitail}{Link} & Textual Entailment \\
\quad 1368 & HealthFact & health\_fact & Sentence Composition & Healthcare & 566.0 & 11.2 & \href{https://huggingface.co/datasets/health_fact}{Link} & Deductive Reasoning \\
\quad 1364 & HANS & hans & Sentence Composition & Movie Reviews & 8.6 & 6.0 & \href{https://arxiv.org/abs/1902.01007}{Link} &  \\
\midrule
\multicolumn{9}{l}{\textbf{Textual Entailment} - Morphology (\textcolor{forestgreen}{+17.5\% $\uparrow$}), Discourse \textcolor{red}{(-6.2\% $\downarrow$})}\\
\quad 640 & e-SNLI & e\_snli & Textual Entailment & Misc. & 23.1 & 1.0 & \href{https://github.com/OanaMariaCamburu/e-SNLI/}{Link} & Textual Entailment \\
\quad 201 & MultiNLI & multinli & Textual Entailment & History, Fiction, Dialogue, Law & 54.9 & 1.0 & \href{https://cims.nyu.edu/~sbowman/multinli/paper.pdf}{Link} & Textual Entailment -> Deductive Reasoning, Commonsense Reasoning \\
\quad 1387 & ANLI R3 & anli & Textual Entailment & Misc. & 67.8 & 1.0 & \href{https://github.com/facebookresearch/anli}{Link} & Textual Entailment, Quantitative Reasoning, Commonsense Reasoning, Causal Reasoning \\
\quad 190 & SNLI & snli & Textual Entailment & Image Captions & 24.2 & 1.0 & \href{https://nlp.stanford.edu/pubs/snli_paper.pdf}{Link} & Textual Entailment \\
\quad 1612 & SICK & sick & Textual Entailment & Image \& Video Captions & 21.1 & 1.0 & \href{https://huggingface.co/datasets/sick}{Link} & Textual Entailment -> Deductive Reasoning \\
\midrule
\multicolumn{9}{l}{\textbf{Named Entity Recognition} - Morphology (\textcolor{forestgreen}{+9.2\% $\uparrow$}), Others \textcolor{red}{(-15.3\% $\downarrow$})}\\
\quad 1480 & JNLPBA & jnlpba\_corpus & Named Entity Recognition & Bioinformatics & 29.5 & 2.5 & \href{https://huggingface.co/datasets/jnlpba}{Link} &  \\
\quad 1486 & ANEM & anem & Named Entity Recognition & Clinical Knowledge & 30.3 & 1.2 & \href{https://github.com/juand-r/entity-recognition-datasets/tree/master/data/AnEM}{Link} &  \\
\quad 959 & E2E & e2e & Named Entity Recognition & Restaurants & 27.1 & 2.1 & \href{https://arxiv.org/abs/1706.09254}{Link} &  \\
\quad 1483 & ChemProt & chemprot\_corpus & Named Entity Recognition & Chemistry & 14.8 & 1.2 & \href{https://paperswithcode.com/dataset/chemprot}{Link} &  \\
\quad 1481 & BC2GM & bc2gm\_corpus & Named Entity Recognition & Bioinformatics & 30.5 & 2.4 & \href{https://metatext.io/datasets/biocreative-ii-gene-mention-recognition-(bc2gm)}{Link} &  \\
\midrule
\multicolumn{9}{l}{\textbf{Text Categorization} - Morphology (\textcolor{forestgreen}{+9.2\% $\uparrow$}), Discourse \textcolor{red}{(-7.8\% $\downarrow$})}\\
\quad 1495 & ADE Corpus V2 & ade\_corpus\_v2 & Text Categorization & Clinical Knowledge, Healthcare & 17.7 & 3.0 & \href{https://github.com/trunghlt/AdverseDrugReaction/tree/master/ADE-Corpus-V2}{Link} &  \\
\quad 1489 & Sarcasm in Twitter & sarcasm\_in\_twitter & Text Categorization & Social Media & 17.7 & 1.0 & \href{https://www.kaggle.com/c/gse002}{Link} & Commonsense Reasoning \\
\quad 1308 & Amazon Reviews & multilingual\_amazon\_reviews & Text Categorization & Reviews & 64.2 & 1.0 & \href{https://huggingface.co/datasets/amazon\_reviews\_multi}{Link} & Reasoning on Social Interactions, Commonsense Reasoning \\
\quad 617 & Amazon Reviews & multilingual\_amazon\_reviews & Text Categorization & Reviews & 68.0 & 1.3 & \href{https://huggingface.co/datasets/amazon\_reviews\_multi}{Link} &  \\
\quad 1541 & AG News & ag\_news & Text Categorization & News & 37.8 & 1.0 & \href{https://huggingface.co/datasets/ag\_news}{Link} &  \\
\midrule
\multicolumn{9}{l}{\textbf{Title Generation} - Morphology (\textcolor{forestgreen}{+12.1\% $\uparrow$}), Lexico-Syntax \textcolor{forestgreen}{(+0.9\% $\uparrow$})}\\
\quad 220 & ROCStories & rocstories & Title Generation & Narrative, Story & 60.0 & 1.0 & \href{https://arxiv.org/pdf/1604.01696.pdf}{Link} & Deductive Reasoning \\
\quad 288 & GigaWord & gigaword & Title Generation & News & 29.9 & 8.8 & \href{https://metatext.io/datasets/gigaword}{Link} &  \\
\quad 619 & OHSUMED & ohsumed & Title Generation & Reviews & 157.5 & 11.2 & \href{https://huggingface.co/datasets/ohsumed}{Link} &  \\
\quad 1540 & PeerRead & peer\_read & Title Generation & Computer Science & 164.9 & 8.4 & \href{https://huggingface.co/datasets/peer\_read}{Link} &  \\
\quad 1659 & BillSum & billsum & Title Generation & Government and Politics & 180.3 & 18.4 & \href{https://huggingface.co/datasets/billsum}{Link} &  \\
\midrule
\multicolumn{9}{l}{\textbf{Information Extraction} - Morphology (\textcolor{forestgreen}{+7.5\% $\uparrow$}), Others \textcolor{red}{(-5.3\% $\downarrow$})}\\
\quad 646 & Winograd WSC & winograd\_wsc & Information Extraction & Narrative & 16.4 & 1.7 & \href{https://huggingface.co/datasets/winograd\_wsc}{Link} &  \\
\quad 1413 & DART & dart & Information Extraction & Wikipedia & 8.9 & 1.7 & \href{https://huggingface.co/datasets/dart}{Link} &  \\
\quad 748 & GLUCOSE & glucose & Information Extraction & Story & 55.4 & 11.5 & \href{https://github.com/ElementalCognition/glucose/}{Link} & Causal Reasoning \\
\quad 1411 & DART & dart & Information Extraction & Wikipedia & 9.1 & 2.2 & \href{https://huggingface.co/datasets/dart}{Link} &  \\
\quad 1506 & Synthetic & synthetic & Information Extraction & Pop & 22.5 & 3.0 & - &  \\
\bottomrule
\end{tabular}
\caption{Overview of the tasks in this study with their average input and output tokens across all examples. We also show the two paraphrase types with the highest and lowest gain or loss across all five sampled tasks within a task category. Table 4/4.}\label{tab:ap_overview_tasks_4}
\end{table*}

\begin{table*}
    \begin{tabular}{l l l r}
    \toprule
    \textbf{Paraphrase Group} & \textbf{Paraphrase Type} & \textbf{Model Output} & \textbf{ROUGE-L} \\
    \midrule
    Morphology & Inflectional & Great Value & \hlgold{1.00} \\
    Extremes & Entailment & Great Value & \hlgold{1.00} \\
    Syntax & Negation Switching & Great Value Cat Food & \hlgreen{0.67} \\
    Morphology & Modal Verb & Great Value Cat Food & \hlgreen{0.67} \\
    Morphology & Derivational & Great Value Cat Food & \hlgreen{0.67} \\
    Discourse & Direct/Indirect Style Altern. & Great Value Cat Food & \hlgreen{0.67} \\
    Lexico-Syntax & Synthetic/Analytic Sub. & Great Value for Cats & \hlgreen{0.67} \\
    Lexico-Syntax & Opp. Pol. Sub. (Habitual) & Great Value for Multiple Cats & \hlgreen{0.57} \\
    Lexicon & Spelling & Great Value for Multiple Cats & \hlgreen{0.57} \\
    Extremes & Non-Paraphrase & Good Value Cat Food & {0.33} \\ \midrule 
    Original & Original & Good Value Cat Food & {0.33} \\ \midrule
    Discourse & Syntax/Discourse Structure & Good Value for Cats & {0.33} \\
    Syntax & Subordination and Nesting & Purrfect Kitten Food & \hlred{0.00} \\
    Others & Addition/Deletion & Kittens loved the food. & \hlred{0.00} \\
    Discourse & Punctuation & Kittens loved the food & \hlred{0.00} \\
    Others & Change of Order & Cats love the food. & \hlred{0.00} \\
    Extremes & Identity & Cats loved the food & \hlred{0.00} \\
    Syntax & Coordination & Kittens Loved the Food & \hlred{0.00} \\
    Syntax & Ellipsis & Good for Kittens & \hlred{0.00} \\
    Syntax & Diathesis Alternation & Positive & \hlred{0.00} \\
    Lexico-Syntax & Opp. Pol. Sub. (Contextual) & Good Quality Cat Food & \hlred{0.00} \\
    Lexico-Syntax & Converse Sub. & Cats Loved Food & \hlred{0.00} \\
    Lexicon & Same Pol. Sub. (Named Ent.) & Kittens loved the food. & \hlred{0.00} \\
    Lexicon & Same Pol. Sub. (Habitual) & Perfect for Multiple Cats & \hlred{0.00} \\
    Lexicon & Same Pol. Sub. (Contextual) & Multiple Cat Food & \hlred{0.00} \\
    Lexicon & Change of Format & Cats loved the food & \hlred{0.00} \\
    Others & Semantic-Based & Cats loved the food & \hlred{0.00} \\
    \bottomrule
    \end{tabular}
    \caption{How we calculate the potential performance gain for paraphrasing prompts using a selected example from the Amazon food review summarization task \cite{mcauley2013amateurs} with LLaMA 3 8B. \hlgreen{Median Gain}. To calculate the median gain (as in \Cref{fig:teaser}), we consider all paraphrases of the prompt that are better than the original prompt (green and yellow). \hlred{Median Loss}. To calculate the median loss (as in \Cref{fig:teaser}), we consider all paraphrases of the prompt that are worse than the original prompt (red). \hlgold{Max}. We calculate the maximum possible gain by aggregating only the best performing paraphrases of the prompts highlighted in yellow.}    \label{tab:ap_potential_gain_explained}
\end{table*}

\begin{table*}[t]
    \centering
    \begin{tabular}{lr}
        \toprule
        \textbf{Paraphrase Type}                             &  \# \textbf{Examples Computed} \\
        \midrule
        \textbf{Morphology}                 &    \textbf{360k} \\
        \quad Derivational Changes                        &     \\
        \quad Inflectional Changes                        &     \\
        \quad Modal Verb Changes                          &     \\
        \midrule
        \textbf{Lexicon}                    &   \textbf{600k} \\
        \quad Spelling changes                            &     \\
        \quad Change of format                            &     \\
        \quad Same Polarity Substitution (contextual)     &    \\
        \quad Same Polarity Substitution (habitual)       &     \\
        \quad Same Polarity Substitution (named ent.)     &     \\
        \midrule
        \textbf{Lexico-syntactic}           &    \textbf{480k} \\
        \quad Converse substitution                       &      \\
        \quad Opposite polarity substitution (contextual) &      \\
        \quad Opposite polarity substitution (habitual)   &       \\
        \quad Synthetic/analytic substitution             &     \\
        \midrule
        \textbf{Syntax}                     &    \textbf{600k} \\
        \quad Coordination changes                        &      \\
        \quad Diathesis alternation                       &     \\
        \quad Ellipsis                                    &      \\
        \quad Negation switching                          &      \\
        \quad Subordination and nesting changes           &     \\
        \midrule
        \textbf{Discourse}                  &    \textbf{360k} \\
        \quad Direct/indirect style alternations          &      \\
        \quad Punctuation changes                         &     \\
        \quad Syntax/discourse structure changes          &     \\ 
        \midrule
        \textbf{Others}                                   &    \textbf{720k}\\
        \quad Addition/Deletion                           &    \\
        \quad Change of order                             &     \\
        \quad Semantic-based                              &     \\ 
        \quad Entailment                                  &      \\
        \quad Identity                                    &    \\
        \quad Non-paraphrase                              &     \\
        \midrule
        {Total}                                     &   \textbf{3,24m} \\
        \bottomrule        
    \end{tabular}
    \caption{An overview of the considered types of paraphrase prompts, categorized into their six main groups. The numbers indicate how many task examples with instructions of that paraphrase type group were run in our experiments (this is a product of all models, tasks, and number of examples per task). The number of paraphrases within a group is equally balanced (i.e., derivational changes have occurred the same amount of time as inflectional changes).}    \label{tab:ap_types_by_group}
\end{table*}

\begin{table*}[t]
    \centering
    \resizebox{\textwidth}{!}{
    \begin{tabular}{lrrrrrr}
    \toprule
    \textbf{Task Family} & \textbf{Morphology} & \textbf{Syntax} & \textbf{Lexicon} & \textbf{Lex.-Syn.} & \textbf{Discourse} & \textbf{Others} \\
    \midrule
    \text{Answerability Classification} & \textcolor{forestgreen}{+7.0\%} & \textcolor{red}{-3.0\%} & \textcolor{forestgreen}{+6.2\%} & \textcolor{forestgreen}{\textbf{+11.8\%}} & +0.9\% & \textcolor{red}{-3.5\%} \\
    \text{Commonsense Classification} & \textcolor{forestgreen}{+14.6\%} & +2.9\% & -2.7\% & -1.9\% & +0.6\% & -1.6\% \\
    \text{Coreference Resolution} & \textcolor{forestgreen}{+8.3\%} & \textcolor{forestgreen}{\textbf{+9.4\%}} & \textcolor{forestgreen}{+4.6\%} & +1.6\% & \textcolor{forestgreen}{+3.3\%} & \textcolor{forestgreen}{\textbf{+25.1\%}} \\
    \text{Dialogue Generation} & \textcolor{forestgreen}{+9.3\%} & \textcolor{forestgreen}{+3.4\%} & \textcolor{forestgreen}{+6.3\%} & \textcolor{forestgreen}{+9.2\%} & \textcolor{red}{-3.9\%} & \textcolor{forestgreen}{+4.6\%} \\
    \text{Fill in The Blank} & \textcolor{forestgreen}{+10.0\%} & \textcolor{red}{-2.0\%} & \textcolor{forestgreen}{+4.7\%} & \textcolor{forestgreen}{+2.7\%} & \textcolor{red}{-5.0\%} & \textcolor{forestgreen}{+18.4\%} \\
    \text{Information Extraction} & \textcolor{forestgreen}{+7.5\%} & -1.7\% & -1.0\% & \textcolor{red}{-5.2\%} & \textcolor{forestgreen}{+6.1\%} & \textcolor{red}{-5.3\%} \\
    \text{Named Entity Recognition} & \textcolor{forestgreen}{+9.2\%} & \textcolor{red}{-8.7\%} & -2.1\% & +2.0\% & \textbf{\textcolor{forestgreen}{+8.7\%}} & \textcolor{red}{-15.3\%} \\
    \text{Program Execution} & +1.94\% & \textcolor{red}{-3.1\%} & -1.5\% & -1.7\% & \textcolor{red}{-4.2\%} & -2.8\% \\
    \text{Question Answering} & \textcolor{forestgreen}{+10.3\%} & \textcolor{forestgreen}{+1.8\%} & \textcolor{red}{-3.0\%} & \textcolor{forestgreen}{+9.9\%} & \textcolor{forestgreen}{+5.0\%} & \textcolor{forestgreen}{+24.8\%} \\
    \text{Question Generation} & \textcolor{forestgreen}{+21.5\%} & \textcolor{forestgreen}{+5.9\%} & +2.5\% & \textcolor{forestgreen}{+5.9\%} & +1.3\% & \textcolor{forestgreen}{+4.1\%} \\
    \text{Question Rewriting} & \textcolor{red}{-4.7\%} & \textcolor{red}{-17.2\%} & \textcolor{red}{-6.6\%} & \textcolor{red}{-5.8\%} & \textcolor{red}{-8.2\%} & \textcolor{red}{-8.6\%} \\
    \text{Question Understanding} & \textcolor{forestgreen}{+15.1\%} & \textcolor{forestgreen}{+3.5\%} & -2.7\% & -2.8\% & \textcolor{red}{-4.1\%} & \textcolor{red}{-3.3\%} \\
    \text{Sentence Composition} & \textcolor{red}{-5.0\%} & \textcolor{forestgreen}{+3.4\%} & -2.9\% & -2.7\% & \textcolor{forestgreen}{+3.8\%} & +2.9\% \\
    \text{Sentiment Analysis} & \textcolor{forestgreen}{+15.6\%} & +1.6\% & +1.7\% & +1.9\% & -2.1\% & -1.7\% \\
    \text{Summarization} & \textcolor{forestgreen}{+6.9\%} & +2.6\% & \textcolor{forestgreen}{\textbf{+10.8\%}} & \textcolor{forestgreen}{+4.8\%} & \textcolor{forestgreen}{+7.8\%} & \textcolor{forestgreen}{+7.5\%} \\
    \text{Text Categorization} & \textcolor{forestgreen}{+9.2\%} & \textcolor{red}{-5.3\%} & \textcolor{red}{-7.0\%} & -1.6\% & \textcolor{red}{-7.8\%} & -2.4\% \\
    \text{Text Completion} & \textcolor{forestgreen}{+11.4\%} & \textcolor{red}{-5.3\%} & \textcolor{red}{-3.0\%} & -2.6\% & \textcolor{red}{-3.1\%} & +1.7\% \\
    \text{Text Matching} & +0.5\% & \textcolor{forestgreen}{+5.4\%} & -1.1\% & +0.3\% & \textcolor{red}{-12.2\%} & -1.9\% \\
    \text{Text Quality Evaluation} & \textcolor{forestgreen}{+7.7\%} & -1.1\% & +1.4\% & +1.2\% & \textcolor{red}{-8.6\%} & +1.4\% \\
    \text{Text to Code} & +2.4\% & -2.1\% & -1.7\% & \textcolor{red}{-5.5\%} & -0.6\% & +0.7\% \\
    \text{Textual Entailment} & \textcolor{forestgreen}{+17.5\%} & \textcolor{red}{-4.2\%} & \textcolor{forestgreen}{+5.2\%} & \textcolor{forestgreen}{+5.3\%} & \textcolor{red}{-6.2\%} & \textcolor{red}{-4.6\%} \\
    \text{Title Generation} & \textcolor{forestgreen}{+12.1\%} & \textcolor{forestgreen}{+5.8\%} & \textcolor{forestgreen}{+7.1\%} & +1.7\% & +0.9\% & \textcolor{forestgreen}{+7.4\%} \\
    \text{Word Semantics} & \textcolor{forestgreen}{+14.9\%} & -0.5\% & -2.7\% & \textcolor{red}{-6.5\%} & \textcolor{forestgreen}{+3.1\%} & +2.6\% \\
    \text{Wrong Candidate Generation} & \textcolor{forestgreen}{\textbf{+26.0\%}} & -1.4\% & \textcolor{forestgreen}{+8.9\%} & \textcolor{forestgreen}{+7.5\%} & -2.9\% & \textcolor{forestgreen}{+5.9\%} \\
    \bottomrule
    \end{tabular}
    }
    \caption{The average downstream task performance gain or loss over the original prompt when paraphrased with a specific type from one of the six groups (columns) of tasks within a certain task family (rows) as an average over \textbf{all models}. We calculate the average across all paraphrase types in one of the six paraphrase groups and across all five tasks within one of the 24 categories. \textbf{Bold} indicates the highest score per column. Small changes between -3\% and +3\% are not colored.}
    \label{tab:ap_task_by_types}
\end{table*}

\begin{table*}[t]
    \centering
    \resizebox{\textwidth}{!}{
\begin{tabular}{lrrrrrr}
\toprule
\textbf{Task Family} & \textbf{Morphology} & \textbf{Syntax} & \textbf{Lexicon} & \textbf{Lex.-Syn.} & \textbf{Discourse} & \textbf{Others} \\
\midrule
\text{Answerability Classification} & \textcolor{red}{-9.0\% $\downarrow$} & -2.1\% & \textbf{\textcolor{forestgreen}{+21.9\% $\uparrow$}} & \textbf{\textcolor{forestgreen}{+57.6\% $\uparrow$}} & 0.0\% & 0.0\% \\
\text{Commonsense Classification} & \textcolor{forestgreen}{+6.4\% $\uparrow$} & 0.0\% & \textcolor{red}{-3.1\% $\downarrow$} & -2.4\% & 0.0\% & 0.0\% \\
\text{Coreference Resolution} & \textcolor{forestgreen}{+6.4\% $\uparrow$} & \textbf{\textcolor{forestgreen}{+35.0\% $\uparrow$}} & \textcolor{forestgreen}{+10.8\% $\uparrow$} & \textcolor{forestgreen}{+18.2\% $\uparrow$} & 0.0\% & \textcolor{forestgreen}{+25.0\% $\uparrow$} \\
\text{Dialogue Generation} & \textcolor{forestgreen}{+8.4\% $\uparrow$} & +1.6\% & \textcolor{forestgreen}{+4.5\% $\uparrow$} & +1.6\% & -2.0\% & \textcolor{forestgreen}{+28.6\% $\uparrow$} \\
\text{Fill in The Blank} & \textcolor{red}{-18.4\% $\downarrow$} & \textcolor{forestgreen}{+4.5\% $\uparrow$} & \textcolor{forestgreen}{+15.4\% $\uparrow$} & \textcolor{forestgreen}{+3.3\% $\uparrow$} & \textcolor{red}{-10.4\% $\downarrow$} & \textbf{\textcolor{forestgreen}{+36.0\% $\uparrow$}} \\
\text{Information Extraction} & \textcolor{forestgreen}{+8.7\% $\uparrow$} & \textcolor{red}{-11.4\% $\downarrow$} & \textcolor{red}{-13.6\% $\downarrow$} & \textcolor{red}{-15.3\% $\downarrow$} & \textcolor{forestgreen}{+17.9\% $\uparrow$} & \textcolor{red}{-20.0\% $\downarrow$} \\
\text{Named Entity Recognition} & \textcolor{forestgreen}{+9.0\% $\uparrow$} & \textcolor{red}{-42.3\% $\downarrow$} & \textcolor{red}{-16.2\% $\downarrow$} & \textcolor{forestgreen}{+20.6\% $\uparrow$} & \textbf{\textcolor{forestgreen}{+49.7\% $\uparrow$}} & \textcolor{red}{-71.6\% $\downarrow$} \\
\text{Program Execution} & \textcolor{forestgreen}{+9.7\% $\uparrow$} & \textcolor{red}{-11.7\% $\downarrow$} & \textcolor{red}{-7.9\% $\downarrow$} & \textcolor{red}{-7.5\% $\downarrow$} & \textcolor{red}{-18.9\% $\downarrow$} & \textcolor{red}{-23.7\% $\downarrow$} \\
\text{Question Answering} & \textcolor{forestgreen}{+7.6\% $\uparrow$} & \textcolor{forestgreen}{+9.6\% $\uparrow$} & \textcolor{red}{-6.6\% $\downarrow$} & \textcolor{forestgreen}{+4.4\% $\uparrow$} & \textcolor{red}{-4.9\% $\downarrow$} & \textcolor{forestgreen}{+13.0\% $\uparrow$} \\
\text{Question Generation} & \textcolor{forestgreen}{+19.4\% $\uparrow$} & \textcolor{red}{-16.4\% $\downarrow$} & \textcolor{red}{-10.8\% $\downarrow$} & \textcolor{red}{-11.8\% $\downarrow$} & \textcolor{red}{-10.7\% $\downarrow$} & \textcolor{red}{-6.5\% $\downarrow$} \\
\text{Question Rewriting} & \textcolor{red}{-8.3\% $\downarrow$} & \textcolor{red}{-9.0\% $\downarrow$} & \textcolor{red}{-5.8\% $\downarrow$} & \textcolor{red}{-8.1\% $\downarrow$} & \textcolor{red}{-26.1\% $\downarrow$} & \textcolor{red}{-15.9\% $\downarrow$} \\
\text{Question Understanding} & \textcolor{red}{-12.3\% $\downarrow$} & \textcolor{red}{-14.6\% $\downarrow$} & \textcolor{red}{-17.5\% $\downarrow$} & \textcolor{red}{-21.3\% $\downarrow$} & \textcolor{red}{-13.2\% $\downarrow$} & \textcolor{red}{-34.8\% $\downarrow$} \\
\text{Sentence Composition} & \textcolor{red}{-23.7\% $\downarrow$} & \textcolor{forestgreen}{+6.3\% $\uparrow$} & \textcolor{forestgreen}{+4.6\% $\uparrow$} & \textcolor{forestgreen}{+4.7\% $\uparrow$} & \textcolor{forestgreen}{+6.4\% $\uparrow$} & +1.7\% \\
\text{Sentiment Analysis} & \textcolor{forestgreen}{+3.5\% $\uparrow$} & 0.0\% & -2.5\% & \textcolor{red}{-5.6\% $\downarrow$} & 0.0\% & 0.0\% \\
\text{Summarization} & \textcolor{forestgreen}{+14.6\% $\uparrow$} & \textcolor{red}{-4.0\% $\downarrow$} & \textcolor{forestgreen}{+7.9\% $\uparrow$} & \textcolor{forestgreen}{+9.5\% $\uparrow$} & \textcolor{red}{-5.3\% $\downarrow$} & \textcolor{forestgreen}{+4.3\% $\uparrow$} \\
\text{Text Categorization} & \textcolor{forestgreen}{+5.3\% $\uparrow$} & \textcolor{red}{-19.5\% $\downarrow$} & \textcolor{red}{-30.8\% $\downarrow$} & \textcolor{red}{-13.2\% $\downarrow$} & \textcolor{red}{-41.2\% $\downarrow$} & 0.0\% \\
\text{Text Completion} & \textcolor{forestgreen}{+20.0\% $\uparrow$} & \textcolor{red}{-33.7\% $\downarrow$} & \textcolor{red}{-30.3\% $\downarrow$} & \textcolor{red}{-18.8\% $\downarrow$} & \textcolor{red}{-17.1\% $\downarrow$} & \textcolor{red}{-10.0\% $\downarrow$} \\
\text{Text Matching} & +3.0\% & \textcolor{forestgreen}{+22.1\% $\uparrow$} & \textcolor{forestgreen}{+5.5\% $\uparrow$} & -1.0\% & \textcolor{red}{-51.3\% $\downarrow$} & \textcolor{red}{-22.7\% $\downarrow$} \\
\text{Text Quality Evaluation} & \textcolor{red}{-5.0\% $\downarrow$} & \textcolor{red}{-3.2\% $\downarrow$} & \textcolor{red}{-4.0\% $\downarrow$} & \textcolor{red}{-7.4\% $\downarrow$} & \textcolor{red}{-65.4\% $\downarrow$} & 0.0\% \\
\text{Text to Code} & \textcolor{forestgreen}{+8.0\% $\uparrow$} & \textcolor{forestgreen}{+3.9\% $\uparrow$} & \textcolor{forestgreen}{+10.0\% $\uparrow$} & -2.6\% & -1.0\% & 0.0\% \\
\text{Textual Entailment} & \textcolor{forestgreen}{+16.3\% $\uparrow$} & \textcolor{red}{-31.1\% $\downarrow$} & \textcolor{forestgreen}{+17.0\% $\uparrow$} & \textcolor{forestgreen}{+14.9\% $\uparrow$} & \textcolor{red}{-35.5\% $\downarrow$} & -1.2\% \\
\text{Title Generation} & \textcolor{forestgreen}{+11.6\% $\uparrow$} & \textcolor{forestgreen}{+9.9\% $\uparrow$} & \textcolor{forestgreen}{+9.3\% $\uparrow$} & \textcolor{red}{-13.1\% $\downarrow$} & \textcolor{red}{-6.3\% $\downarrow$} & -1.8\% \\
\text{Word Semantics} & \textcolor{forestgreen}{+4.9\% $\uparrow$} & \textcolor{red}{-7.8\% $\downarrow$} & \textcolor{red}{-7.7\% $\downarrow$} & \textcolor{red}{-19.6\% $\downarrow$} & \textcolor{forestgreen}{+9.8\% $\uparrow$} & -0.1\% \\
\text{Wrong Candidate Generation} & \textbf{\textcolor{forestgreen}{+24.5\% $\uparrow$}} & \textcolor{forestgreen}{+6.4\% $\uparrow$} & \textcolor{forestgreen}{+8.4\% $\uparrow$} & \textcolor{forestgreen}{+9.6\% $\uparrow$} & \textcolor{forestgreen}{+8.9\% $\uparrow$} & \textcolor{forestgreen}{+5.1\% $\uparrow$} \\
\bottomrule
\end{tabular}
}
    \caption{The average downstream task performance gain or loss over the original prompt when paraphrased with a specific type from one of the six groups (columns) of tasks within a certain task family (rows) for \textbf{Mixtral 8x7B Instruct (47B)}. We calculate the average across all paraphrase types in one of the six paraphrase groups and across all five tasks within one of the 24 categories. \textbf{Bold} indicates the highest score per column. Small changes between -3\% and +3\% are not colored.}
    \label{tab:ap_task_by_types_mixtral}
\end{table*}

\begin{table*}[t]
    \centering
    \resizebox{\textwidth}{!}{
    \begin{tabular}{lrrrrrr}
\toprule
\textbf{Task Family} & \textbf{Morphology} & \textbf{Syntax} & \textbf{Lexicon} & \textbf{Lex.-Syn.} & \textbf{Discourse} & \textbf{Others} \\
\midrule
\text{Answerability Classification} & \textcolor{forestgreen}{+34.7\% $\uparrow$} & \textcolor{red}{-10.5\% $\downarrow$} & \textcolor{forestgreen}{+9.2\% $\uparrow$} & \textcolor{forestgreen}{+5.0\% $\uparrow$} & \textcolor{forestgreen}{+4.6\% $\uparrow$} & \textcolor{red}{-17.6\% $\downarrow$} \\
\text{Commonsense Classification} & \textcolor{forestgreen}{+44.9\% $\uparrow$} & \textcolor{forestgreen}{+14.3\% $\uparrow$} & \textcolor{red}{-10.4\% $\downarrow$} & \textcolor{forestgreen}{+11.2\% $\uparrow$} & +2.8\% & \textcolor{red}{-7.9\% $\downarrow$} \\
\text{Coreference Resolution} & \textcolor{forestgreen}{+23.1\% $\uparrow$} & \textcolor{red}{-10.0\% $\downarrow$} & \textcolor{forestgreen}{+11.1\% $\uparrow$} & \textcolor{red}{-8.9\% $\downarrow$} & \textcolor{red}{-4.8\% $\downarrow$} & +0.4\% \\
\text{Dialogue Generation} & \textcolor{forestgreen}{+15.5\% $\uparrow$} & \textcolor{red}{-11.3\% $\downarrow$} & \textcolor{forestgreen}{+11.4\% $\uparrow$} & \textcolor{forestgreen}{+11.8\% $\uparrow$} & +0.9\% & \textcolor{red}{-8.9\% $\downarrow$} \\
\text{Fill in The Blank} & \textcolor{forestgreen}{+30.7\% $\uparrow$} & \textcolor{red}{-11.3\% $\downarrow$} & \textcolor{forestgreen}{+12.5\% $\uparrow$} & \textcolor{red}{-7.9\% $\downarrow$} & \textcolor{red}{-16.7\% $\downarrow$} & \textcolor{forestgreen}{+17.6\% $\uparrow$} \\
\text{Information Extraction} & \textcolor{forestgreen}{+17.8\% $\uparrow$} & \textcolor{red}{-4.7\% $\downarrow$} & \textcolor{forestgreen}{+8.2\% $\uparrow$} & \textcolor{red}{-8.3\% $\downarrow$} & \textcolor{forestgreen}{+12.6\% $\uparrow$} & \textcolor{red}{-6.5\% $\downarrow$} \\
\text{Named Entity Recognition} & \textcolor{forestgreen}{+17.2\% $\uparrow$} & \textcolor{red}{-5.2\% $\downarrow$} & \textcolor{forestgreen}{+9.0\% $\uparrow$} & \textcolor{red}{-9.5\% $\downarrow$} & \textcolor{red}{-5.7\% $\downarrow$} & \textcolor{red}{-5.1\% $\downarrow$} \\
\text{Program Execution} & \textcolor{forestgreen}{+22.6\% $\uparrow$} & \textcolor{red}{-7.1\% $\downarrow$} & \textcolor{forestgreen}{+8.3\% $\uparrow$} & \textcolor{red}{-9.1\% $\downarrow$} & \textcolor{forestgreen}{+4.7\% $\uparrow$} & \textcolor{forestgreen}{+9.6\% $\uparrow$} \\
\text{Question Answering} & \textcolor{forestgreen}{+17.6\% $\uparrow$} & \textcolor{forestgreen}{+4.7\% $\uparrow$} & \textbf{\textcolor{forestgreen}{+15.9\% $\uparrow$}} & \textcolor{forestgreen}{+8.0\% $\uparrow$} & \textcolor{forestgreen}{+12.4\% $\uparrow$} & \textcolor{red}{-8.3\% $\downarrow$} \\
\text{Question Generation} & \textcolor{forestgreen}{+18.1\% $\uparrow$} & \textcolor{forestgreen}{+11.1\% $\uparrow$} & \textcolor{forestgreen}{+14.0\% $\uparrow$} & \textbf{\textcolor{forestgreen}{+16.6\% $\uparrow$} }& \textcolor{red}{-8.7\% $\downarrow$} & \textbf{\textcolor{forestgreen}{+30.4\% $\uparrow$}} \\
\text{Question Rewriting} & \textcolor{red}{-13.6\% $\downarrow$} & \textcolor{red}{-4.0\% $\downarrow$} & \textcolor{red}{-12.3\% $\downarrow$} & \textcolor{forestgreen}{+6.8\% $\uparrow$} & \textcolor{red}{-5.9\% $\downarrow$} & \textcolor{red}{-7.5\% $\downarrow$} \\
\text{Question Understanding} & \textcolor{forestgreen}{+35.3\% $\uparrow$} & \textbf{\textcolor{forestgreen}{+17.7\% $\uparrow$}} & \textcolor{red}{-11.3\% $\downarrow$} & \textcolor{red}{-10.6\% $\downarrow$} & \textcolor{red}{-9.6\% $\downarrow$} & -2.3\% \\
\text{Sentence Composition} & \textcolor{forestgreen}{+14.7\% $\uparrow$} & \textcolor{red}{-6.2\% $\downarrow$} & \textcolor{red}{-9.4\% $\downarrow$} & \textcolor{forestgreen}{+6.5\% $\uparrow$} & \textcolor{forestgreen}{+6.8\% $\uparrow$} & \textcolor{red}{-4.5\% $\downarrow$} \\
\text{Sentiment Analysis} & \textcolor{forestgreen}{+30.6\% $\uparrow$} & \textcolor{forestgreen}{+8.1\% $\uparrow$} & \textcolor{forestgreen}{+10.0\% $\uparrow$} & \textcolor{forestgreen}{+10.1\% $\uparrow$} & \textcolor{red}{-9.5\% $\downarrow$} & \textcolor{red}{-8.5\% $\downarrow$} \\
\text{Summarization} & \textcolor{red}{-16.1\% $\downarrow$} & \textcolor{red}{-9.9\% $\downarrow$} & \textcolor{forestgreen}{+11.3\% $\uparrow$} & \textcolor{forestgreen}{+10.1\% $\uparrow$} & \textcolor{forestgreen}{+16.9\% $\uparrow$} & \textcolor{forestgreen}{+10.9\% $\uparrow$} \\
\text{Text Categorization} & \textcolor{forestgreen}{+26.3\% $\uparrow$} & \textcolor{red}{-10.9\% $\downarrow$} & \textcolor{red}{-9.5\% $\downarrow$} & \textcolor{forestgreen}{+8.0\% $\uparrow$} & \textcolor{red}{-8.0\% $\downarrow$} & \textcolor{red}{-12.1\% $\downarrow$} \\
\text{Text Completion} & \textcolor{forestgreen}{+24.0\% $\uparrow$} & \textcolor{forestgreen}{+11.7\% $\uparrow$} & \textcolor{forestgreen}{+14.3\% $\uparrow$} & \textcolor{forestgreen}{+10.4\% $\uparrow$} & \textcolor{red}{-4.5\% $\downarrow$} & \textcolor{forestgreen}{+8.8\% $\uparrow$} \\
\text{Text Matching} & \textcolor{red}{-10.5\% $\downarrow$} & \textcolor{red}{-4.4\% $\downarrow$} & \textcolor{red}{-6.8\% $\downarrow$} & \textcolor{forestgreen}{+4.1\% $\uparrow$} & \textcolor{red}{-9.9\% $\downarrow$} & \textcolor{forestgreen}{+13.8\% $\uparrow$} \\
\text{Text Quality Evaluation} & \textcolor{forestgreen}{+28.2\% $\uparrow$} & \textcolor{red}{-7.3\% $\downarrow$} & \textcolor{forestgreen}{+11.6\% $\uparrow$} & \textcolor{forestgreen}{+6.7\% $\uparrow$} & \textbf{\textcolor{forestgreen}{+22.5\% $\uparrow$}} & \textcolor{forestgreen}{+7.1\% $\uparrow$} \\
\text{Text to Code} & \textcolor{red}{-11.6\% $\downarrow$} & \textcolor{red}{-14.3\% $\downarrow$} & \textcolor{red}{-11.4\% $\downarrow$} & \textcolor{red}{-19.2\% $\downarrow$} & \textcolor{forestgreen}{+10.8\% $\uparrow$} & +1.2\% \\
\text{Textual Entailment} & \textbf{\textcolor{forestgreen}{+43.2\% $\uparrow$}} & \textcolor{forestgreen}{+9.1\% $\uparrow$} & \textcolor{forestgreen}{+13.8\% $\uparrow$} & \textcolor{forestgreen}{+11.8\% $\uparrow$} & \textcolor{forestgreen}{+4.9\% $\uparrow$} & \textcolor{red}{-23.3\% $\downarrow$} \\
\text{Title Generation} & \textcolor{forestgreen}{+22.3\% $\uparrow$} & \textcolor{forestgreen}{+9.6\% $\uparrow$} & \textcolor{forestgreen}{+12.0\% $\uparrow$} & \textcolor{forestgreen}{+7.7\% $\uparrow$} & \textcolor{red}{-9.4\% $\downarrow$} & \textcolor{red}{-7.3\% $\downarrow$} \\
\text{Word Semantics} & \textcolor{forestgreen}{+38.5\% $\uparrow$} & \textcolor{forestgreen}{+5.9\% $\uparrow$} & \textcolor{red}{-9.0\% $\downarrow$} & \textcolor{red}{-12.9\% $\downarrow$} & +1.7\% & \textcolor{forestgreen}{+8.3\% $\uparrow$} \\
\text{Wrong Candidate Generation} & \textcolor{forestgreen}{+18.9\% $\uparrow$} & \textcolor{red}{-7.8\% $\downarrow$} & \textcolor{forestgreen}{+9.6\% $\uparrow$} & \textcolor{forestgreen}{+8.2\% $\uparrow$} & \textcolor{red}{-6.1\% $\downarrow$} & \textcolor{forestgreen}{+6.7\% $\uparrow$} \\
\bottomrule
\end{tabular}
}
    \caption{The average downstream task performance gain or loss over the original prompt when paraphrased with a specific type from one of the six groups (columns) of tasks within a certain task family (rows) for \textbf{Gemma 7B Instruct (7B)}. We calculate the average across all paraphrase types in one of the six paraphrase groups and across all five tasks within one of the 24 categories. \textbf{Bold} indicatess the highest score per paraphrase group (column). Small changes between -3\% and +3\% are not colored.}
    \label{tab:ap_task_by_types_gemma}
\end{table*}

\begin{table*}[t]
    \centering
    \resizebox{\textwidth}{!}{
\begin{tabular}{lrrrrrr}
\toprule
\textbf{Task Family} & \textbf{Morphology} & \textbf{Syntax} & \textbf{Lexicon} & \textbf{Lex.-Syn.} & \textbf{Discourse} & \textbf{Others} \\
\midrule
\text{Answerability Classification} & \textcolor{forestgreen}{+3.2\% $\uparrow$} & -2.6\% & -0.1\% & -1.7\% & 0.0\% & 0.0\% \\
\text{Commonsense Classification} & \textcolor{forestgreen}{+10.4\% $\uparrow$} & 0.0\% & 0.0\% & 0.0\% & 0.0\% & 0.0\% \\
\text{Coreference Resolution} & \textcolor{forestgreen}{+5.8\% $\uparrow$} & 0.0\% & -1.3\% & 0.0\% & 0.0\% & 0.0\% \\
\text{Dialogue Generation} & \textcolor{forestgreen}{+11.2\% $\uparrow$} & \textbf{\textcolor{forestgreen}{+19.4\% $\uparrow$}} & \textcolor{forestgreen}{+5.6\% $\uparrow$} & \textbf{\textcolor{forestgreen}{+18.4\% $\uparrow$}} & \textcolor{red}{-8.5\% $\downarrow$} & 0.0\% \\
\text{Fill in The Blank} & \textcolor{forestgreen}{+13.3\% $\uparrow$} & \textcolor{red}{-3.6\% $\downarrow$} & 0.0\% & 0.0\% & 0.0\% & 0.0\% \\
\text{Information Extraction} & \textcolor{forestgreen}{+5.5\% $\uparrow$} & \textcolor{forestgreen}{+6.6\% $\uparrow$} & \textcolor{forestgreen}{+3.9\% $\uparrow$} & -2.8\% & -0.4\% & 0.0\% \\
\text{Named Entity Recognition} & \textcolor{forestgreen}{+11.5\% $\uparrow$} & 0.0\% & \textcolor{red}{-3.3\% $\downarrow$} & -0.7\% & 0.0\% & 0.0\% \\
\text{Program Execution} & \textcolor{red}{-9.9\% $\downarrow$} & \textcolor{red}{-4.7\% $\downarrow$} & \textcolor{red}{-3.5\% $\downarrow$} & \textcolor{red}{-3.2\% $\downarrow$} & \textcolor{red}{-5.9\% $\downarrow$} & 0.0\% \\
\text{Question Answering} & \textcolor{forestgreen}{+10.6\% $\uparrow$} & \textcolor{red}{-10.7\% $\downarrow$} & \textcolor{red}{-6.8\% $\downarrow$} & \textcolor{forestgreen}{+16.2\% $\uparrow$} & \textcolor{forestgreen}{+8.3\% $\uparrow$} & \textbf{\textcolor{forestgreen}{+30.0\% $\uparrow$}} \\
\text{Question Generation} & \textbf{\textcolor{forestgreen}{+38.3\% $\uparrow$}} & \textcolor{forestgreen}{+13.4\% $\uparrow$} & \textcolor{forestgreen}{+13.1\% $\uparrow$} & \textcolor{red}{-8.9\% $\downarrow$} & \textcolor{forestgreen}{+7.3\% $\uparrow$} & \textcolor{forestgreen}{+12.7\% $\uparrow$} \\
\text{Question Rewriting} & \textcolor{red}{-6.5\% $\downarrow$} & \textcolor{red}{-9.1\% $\downarrow$} & \textcolor{red}{-6.9\% $\downarrow$} & \textcolor{red}{-5.9\% $\downarrow$} & -2.4\% & \textcolor{red}{-4.2\% $\downarrow$} \\
\text{Question Understanding} & \textcolor{forestgreen}{+10.9\% $\uparrow$} & 0.0\% & 0.0\% & 0.0\% & 0.0\% & 0.0\% \\
\text{Sentence Composition} & \textcolor{red}{-14.5\% $\downarrow$} & +2.1\% & \textcolor{forestgreen}{+4.8\% $\uparrow$} & \textcolor{red}{-4.8\% $\downarrow$} & -0.7\% & \textcolor{forestgreen}{+3.8\% $\uparrow$} \\
\text{Sentiment Analysis} & \textcolor{forestgreen}{+7.5\% $\uparrow$} & 0.0\% & -1.2\% & 0.0\% & 0.0\% & 0.0\% \\
\text{Summarization} & \textcolor{forestgreen}{+23.6\% $\uparrow$} & \textcolor{forestgreen}{+6.6\% $\uparrow$} & \textbf{\textcolor{forestgreen}{+15.5\% $\uparrow$}} & \textcolor{red}{-9.2\% $\downarrow$} & \textbf{\textcolor{forestgreen}{+15.7\% $\uparrow$}} & \textcolor{red}{-11.1\% $\downarrow$} \\
\text{Text Categorization} & \textcolor{forestgreen}{+3.6\% $\uparrow$} & 0.0\% & -0.2\% & 0.0\% & 0.0\% & 0.0\% \\
\text{Text Completion} & \textcolor{forestgreen}{+5.5\% $\uparrow$} & +1.6\% & +2.7\% & \textcolor{forestgreen}{+3.3\% $\uparrow$} & -0.4\% & -0.1\% \\
\text{Text Matching} & \textcolor{forestgreen}{+3.5\% $\uparrow$} & 0.0\% & 0.0\% & -0.2\% & 0.0\% & 0.0\% \\
\text{Text Quality Evaluation} & \textcolor{forestgreen}{+6.0\% $\uparrow$} & 0.0\% & -0.4\% & 0.0\% & 0.0\% & 0.0\% \\
\text{Text to Code} & \textcolor{forestgreen}{+4.8\% $\uparrow$} & -0.7\% & -1.4\% & -1.0\% & \textcolor{red}{-4.3\% $\downarrow$} & 0.0\% \\
\text{Textual Entailment} & \textcolor{forestgreen}{+11.0\% $\uparrow$} & 0.0\% & 0.0\% & 0.0\% & 0.0\% & 0.0\% \\
\text{Title Generation} & \textcolor{forestgreen}{+12.8\% $\uparrow$} & \textcolor{forestgreen}{+8.2\% $\uparrow$} & \textcolor{forestgreen}{+14.2\% $\uparrow$} & \textcolor{forestgreen}{+12.0\% $\uparrow$} & \textcolor{forestgreen}{+5.6\% $\uparrow$} & \textcolor{forestgreen}{+10.5\% $\uparrow$} \\
\text{Word Semantics} & \textcolor{forestgreen}{+14.1\% $\uparrow$} & 0.0\% & 0.0\% & 0.0\% & 0.0\% & 0.0\% \\
\text{Wrong Candidate Generation} & \textcolor{forestgreen}{+37.8\% $\uparrow$} & \textcolor{red}{-3.1\% $\downarrow$} & \textcolor{forestgreen}{+10.2\% $\uparrow$} & \textcolor{forestgreen}{+10.9\% $\uparrow$} & \textcolor{red}{-11.7\% $\downarrow$} & -2.0\% \\
\bottomrule
\end{tabular}
}
    \caption{The average downstream task performance gain or loss over the original prompt when paraphrased with a specific type from one of the six groups (columns) of tasks within a certain task family (rows) for \textbf{Command R+ (104B)}. We calculate the average across all paraphrase types in one of the six paraphrase groups and across all five tasks within one of the 24 categories. \textbf{Bold} indicatess the highest score per paraphrase group (column). Small changes between -3\% and +3\% are not colored.}
    \label{tab:ap_task_by_types_command_r_plus}
\end{table*}

\begin{table*}[t]
    \centering
    \resizebox{\textwidth}{!}{
\begin{tabular}{lrrrrrr}
\toprule
\textbf{Task Family} & \textbf{Morphology} & \textbf{Syntax} & \textbf{Lexicon} & \textbf{Lex.-Syn.} & \textbf{Discourse} & \textbf{Others} \\
\midrule
\text{Answerability Classification} & \textcolor{forestgreen}{+4.2\% $\uparrow$} & 0.0\% & 0.0\% & 0.0\% & 0.0\% & 0.0\% \\
\text{Commonsense Classification} & \textcolor{forestgreen}{+7.8\% $\uparrow$} & 0.0\% & 0.0\% & 0.0\% & 0.0\% & 0.0\% \\
\text{Coreference Resolution} & \textcolor{forestgreen}{+8.0\% $\uparrow$} & 0.0\% & +2.2\% & -0.7\% & 0.0\% & 0.0\% \\
\text{Dialogue Generation} & \textcolor{forestgreen}{+6.9\% $\uparrow$} & \textcolor{forestgreen}{+4.9\% $\uparrow$} & \textcolor{forestgreen}{+6.0\% $\uparrow$} & \textcolor{forestgreen}{+7.7\% $\uparrow$} & \textcolor{red}{-9.6\% $\downarrow$} & \textcolor{forestgreen}{+3.3\% $\uparrow$} \\
\text{Fill in The Blank} & \textcolor{forestgreen}{+19.7\% $\uparrow$} & 0.0\% & 0.0\% & 0.0\% & 0.0\% & 0.0\% \\
\text{Information Extraction} & \textcolor{forestgreen}{+3.1\% $\uparrow$} & -0.6\% & -2.4\% & +1.2\% & -0.5\% & 0.0\% \\
\text{Named Entity Recognition} & \textcolor{forestgreen}{+5.8\% $\uparrow$} & -1.4\% & +0.6\% & -0.2\% & 0.0\% & 0.0\% \\
\text{Program Execution} & \textcolor{red}{-8.5\% $\downarrow$} & \textcolor{forestgreen}{+5.5\% $\uparrow$} & \textcolor{forestgreen}{+7.3\% $\uparrow$} & \textcolor{forestgreen}{+6.1\% $\uparrow$} & -1.1\% & 0.0\% \\
\text{Question Answering} & \textcolor{forestgreen}{+11.5\% $\uparrow$} & \textcolor{forestgreen}{+5.4\% $\uparrow$} & \textcolor{forestgreen}{+9.5\% $\uparrow$} & \textcolor{forestgreen}{+10.7\% $\uparrow$} & -0.6\% & \textbf{\textcolor{forestgreen}{+37.7\% $\uparrow$}} \\
\text{Question Generation} & \textcolor{forestgreen}{+19.4\% $\uparrow$} & \textcolor{forestgreen}{+13.6\% $\uparrow$} & \textbf{\textcolor{forestgreen}{+14.1\% $\uparrow$}} & \textbf{\textcolor{forestgreen}{+15.7\% $\uparrow$}} & \textcolor{forestgreen}{+6.5\% $\uparrow$} & \textcolor{red}{-15.0\% $\downarrow$} \\
\text{Question Rewriting} & \textcolor{forestgreen}{+9.1\% $\uparrow$} & \textcolor{red}{-10.0\% $\downarrow$} & \textcolor{red}{-9.2\% $\downarrow$} & \textcolor{red}{-6.6\% $\downarrow$} & \textcolor{red}{-7.5\% $\downarrow$} & \textcolor{red}{-8.2\% $\downarrow$} \\
\text{Question Understanding} & \textcolor{forestgreen}{+9.6\% $\uparrow$} & 0.0\% & 0.0\% & 0.0\% & 0.0\% & 0.0\% \\
\text{Sentence Composition} & \textcolor{forestgreen}{+7.3\% $\uparrow$} & \textcolor{forestgreen}{+10.3\% $\uparrow$} & \textcolor{red}{-7.9\% $\downarrow$} & \textcolor{red}{-8.5\% $\downarrow$} & \textbf{\textcolor{forestgreen}{+11.0\% $\uparrow$}} & \textcolor{red}{-6.8\% $\downarrow$} \\
\text{Sentiment Analysis} & \textcolor{forestgreen}{+4.2\% $\uparrow$} & 0.0\% & 0.0\% & 0.0\% & 0.0\% & 0.0\% \\
\text{Summarization} & \textcolor{red}{-16.4\% $\downarrow$} & \textcolor{forestgreen}{+10.4\% $\uparrow$} & \textcolor{forestgreen}{+11.2\% $\uparrow$} & \textcolor{forestgreen}{+12.7\% $\uparrow$} & \textcolor{forestgreen}{+9.5\% $\uparrow$} & \textcolor{forestgreen}{+30.9\% $\uparrow$} \\
\text{Text Categorization} & \textcolor{forestgreen}{+5.0\% $\uparrow$} & 0.0\% & -0.1\% & 0.0\% & 0.0\% & 0.0\% \\
\text{Text Completion} & \textcolor{forestgreen}{+4.9\% $\uparrow$} & \textbf{\textcolor{forestgreen}{+15.9\% $\uparrow$}} & \textcolor{red}{-4.3\% $\downarrow$} & \textcolor{red}{-7.1\% $\downarrow$} & 0.0\% & 0.0\% \\
\text{Text Matching} & \textcolor{forestgreen}{+4.1\% $\uparrow$} & 0.0\% & -0.3\% & 0.0\% & 0.0\% & 0.0\% \\
\text{Text Quality Evaluation} & \textcolor{forestgreen}{+5.3\% $\uparrow$} & 0.0\% & 0.0\% & 0.0\% & 0.0\% & 0.0\% \\
\text{Text to Code} & \textcolor{forestgreen}{+10.1\% $\uparrow$} & \textcolor{red}{-4.1\% $\downarrow$} & \textcolor{red}{-3.8\% $\downarrow$} & \textcolor{red}{-3.2\% $\downarrow$} & \textcolor{red}{-6.5\% $\downarrow$} & \textcolor{forestgreen}{+4.5\% $\uparrow$} \\
\text{Textual Entailment} & \textcolor{forestgreen}{+13.4\% $\uparrow$} & 0.0\% & 0.0\% & 0.0\% & 0.0\% & 0.0\% \\
\text{Title Generation} & \textcolor{forestgreen}{+10.9\% $\uparrow$} & \textcolor{forestgreen}{+13.1\% $\uparrow$} & \textcolor{red}{-9.0\% $\downarrow$} & \textcolor{red}{-14.0\% $\downarrow$} & \textcolor{forestgreen}{+3.6\% $\uparrow$} & \textcolor{forestgreen}{+34.5\% $\uparrow$} \\
\text{Word Semantics} & \textcolor{forestgreen}{+9.6\% $\uparrow$} & 0.0\% & 0.0\% & 0.0\% & 0.0\% & 0.0\% \\
\text{Wrong Candidate Generation} & \textbf{\textcolor{forestgreen}{+22.1\% $\uparrow$} }& \textcolor{red}{-6.6\% $\downarrow$} & \textcolor{forestgreen}{+6.1\% $\uparrow$} & \textcolor{forestgreen}{+12.4\% $\uparrow$} & \textcolor{red}{-3.1\% $\downarrow$} & -1.7\% \\
\bottomrule
\end{tabular}
}
    \caption{The average downstream task performance gain or loss over the original prompt when paraphrased with a specific type from one of the six groups (columns) of tasks within a certain task family (rows) for \textbf{LLaMA 3 Instruct (8B)}. We calculate the average across all paraphrase types in one of the six paraphrase groups and across all five tasks within one of the 24 categories. \textbf{Bold} indicatess the highest score per paraphrase group (column). Small changes between -3\% and +3\% are not colored.}
    \label{tab:ap_task_by_types_llama_8b}
\end{table*}

\begin{table*}[t]
    \centering
    \resizebox{\textwidth}{!}{
\begin{tabular}{lrrrrrr}
\toprule
\textbf{Task Family} & \textbf{Morphology} & \textbf{Syntax} & \textbf{Lexicon} & \textbf{Lex.-Syn.} & \textbf{Discourse} & \textbf{Others} \\
\midrule
\text{Answerability Classification} & +1.7\% & 0.0\% & 0.0\% & 0.0\% & 0.0\% & 0.0\% \\
\text{Commonsense Classification} & \textcolor{forestgreen}{+3.6\% $\uparrow$} & 0.0\% & 0.0\% & 0.0\% & 0.0\% & 0.0\% \\
\text{Coreference Resolution} & +1.1\% & +2.2\% & 0.0\% & 0.0\% & 0.0\% & 0.0\% \\
\text{Dialogue Generation} & \textcolor{forestgreen}{+4.7\% $\uparrow$} & +2.2\% & \textcolor{forestgreen}{+4.1\% $\uparrow$} & +0.5\% & +0.6\% & \textcolor{forestgreen}{+3.0\% $\uparrow$} \\
\text{Fill in The Blank} & \textcolor{forestgreen}{+4.8\% $\uparrow$} & 0.0\% & 0.0\% & 0.0\% & 0.0\% & 0.0\% \\
\text{Information Extraction} & +2.2\% & \textcolor{red}{-8.7\% $\downarrow$} & \textcolor{red}{-3.3\% $\downarrow$} & -0.8\% & +0.8\% & \textcolor{forestgreen}{+3.4\% $\uparrow$} \\
\text{Named Entity Recognition} & \textcolor{forestgreen}{+3.2\% $\uparrow$} & 0.0\% & -0.5\% & 0.0\% & 0.0\% & 0.0\% \\
\text{Program Execution} & \textcolor{red}{-6.0\% $\downarrow$} & +1.5\% & \textcolor{red}{-4.6\% $\downarrow$} & \textcolor{forestgreen}{+5.1\% $\uparrow$} & 0.0\% & 0.0\% \\
\text{Question Answering} & \textcolor{forestgreen}{+4.3\% $\uparrow$} & \textcolor{forestgreen}{+5.6\% $\uparrow$} & \textcolor{red}{-7.1\% $\downarrow$} & \textcolor{forestgreen}{+10.3\% $\uparrow$} & \textcolor{forestgreen}{+9.9\% $\uparrow$} & \textbf{\textcolor{forestgreen}{+30.4\% $\uparrow$}} \\
\text{Question Generation} & \textcolor{forestgreen}{+12.3\% $\uparrow$} & \textcolor{forestgreen}{+7.8\% $\uparrow$} & \textcolor{red}{-8.1\% $\downarrow$} & \textcolor{forestgreen}{+6.5\% $\uparrow$} & \textcolor{forestgreen}{+10.2\% $\uparrow$} & -3.0\% \\
\text{Question Rewriting} & \textcolor{red}{-4.4\% $\downarrow$} & \textcolor{red}{-53.7\% $\downarrow$} & \textcolor{red}{-5.7\% $\downarrow$} & \textcolor{red}{-13.4\% $\downarrow$} & \textcolor{red}{-4.8\% $\downarrow$} & \textcolor{red}{-4.6\% $\downarrow$} \\
\text{Question Understanding} & \textcolor{forestgreen}{+9.3\% $\uparrow$} & 0.0\% & 0.0\% & 0.0\% & 0.0\% & 0.0\% \\
\text{Sentence Composition} & \textcolor{red}{-8.6\% $\downarrow$} & +2.6\% & +2.8\% & -0.9\% & \textcolor{red}{-8.7\% $\downarrow$} & \textcolor{forestgreen}{+10.6\% $\uparrow$} \\
\text{Sentiment Analysis} & +2.2\% & 0.0\% & 0.0\% & 0.0\% & 0.0\% & 0.0\% \\
\text{Summarization} & \textcolor{forestgreen}{+11.7\% $\uparrow$} & \textbf{\textcolor{forestgreen}{+9.9\% $\uparrow$}} & \textbf{\textcolor{forestgreen}{+8.1\% $\uparrow$}} & \textcolor{forestgreen}{+7.7\% $\uparrow$} & -1.8\% & \textcolor{forestgreen}{+3.6\% $\uparrow$} \\
\text{Text Categorization} & +1.5\% & \textcolor{red}{-6.3\% $\downarrow$} & -0.8\% & \textbf{\textcolor{forestgreen}{+20.0\% $\uparrow$}} & 0.0\% & 0.0\% \\
\text{Text Completion} & +2.7\% & -1.1\% & \textcolor{red}{-4.4\% $\downarrow$} & -0.2\% & \textbf{\textcolor{forestgreen}{+15.2\% $\uparrow$}} & -0.2\% \\
\text{Text Matching} & +1.0\% & 0.0\% & 0.0\% & 0.0\% & 0.0\% & 0.0\% \\
\text{Text Quality Evaluation} & \textcolor{forestgreen}{+4.1\% $\uparrow$} & 0.0\% & 0.0\% & 0.0\% & 0.0\% & 0.0\% \\
\text{Text to Code} & +1.9\% & \textcolor{red}{-4.9\% $\downarrow$} & -2.0\% & -2.4\% & 0.0\% & -1.4\% \\
\text{Textual Entailment} & \textcolor{forestgreen}{+3.8\% $\uparrow$} & 0.0\% & 0.0\% & 0.0\% & 0.0\% & 0.0\% \\
\text{Title Generation} & \textcolor{forestgreen}{+7.4\% $\uparrow$} & \textcolor{forestgreen}{+4.1\% $\uparrow$} & \textcolor{forestgreen}{+7.3\% $\uparrow$} & \textcolor{red}{-7.5\% $\downarrow$} & \textcolor{forestgreen}{+5.0\% $\uparrow$} & 0.0\% \\
\text{Word Semantics} & +2.5\% & 0.0\% & 0.0\% & 0.0\% & 0.0\% & 0.0\% \\
\text{Wrong Candidate Generation} & \textbf{\textcolor{forestgreen}{+26.8\% $\uparrow$}} & \textcolor{red}{-6.6\% $\downarrow$} & \textcolor{forestgreen}{+8.0\% $\uparrow$} & \textcolor{red}{-4.7\% $\downarrow$} & +1.6\% & \textcolor{forestgreen}{+21.4\% $\uparrow$} \\
\bottomrule
\end{tabular}
}
    \caption{The average downstream task performance gain or loss over the original prompt when paraphrased with a specific type from one of the six groups (columns) of tasks within a certain task family (rows) for \textbf{LLaMA 3 Instruct (70B)}. We calculate the average across all paraphrase types in one of the six paraphrase groups and across all five tasks within one of the 24 categories. \textbf{Bold} indicatess the highest score per paraphrase group (column). Small changes between -3\% and +3\% are not colored.}
    \label{tab:ap_task_by_types_llama_70b}
\end{table*}

\end{document}